\documentclass{article}

\usepackage{microtype}
\usepackage{graphicx}
\usepackage{subfigure}
\usepackage{booktabs} % for professional tables

\usepackage{hyperref}

\usepackage[ruled,vlined,linesnumbered,noend]{algorithm2e}
\usepackage[accepted]{icml2025}

\usepackage{amsmath}
\usepackage{amssymb}
\usepackage{mathtools}
\usepackage{amsthm}

\usepackage[capitalize,noabbrev]{cleveref}

\usepackage[textsize=tiny]{todonotes}

\usepackage{amsmath,amsfonts,bm}

\def\1{\bm{1}}

\def\eps{{\epsilon}}

\def\rv{{\textnormal{v}}}

\def\rvv{{\mathbf{v}}}
\def\rvw{{\mathbf{w}}}
\def\rvx{{\mathbf{x}}}
\def\rvy{{\mathbf{y}}}

\def\rmI{{\mathbf{I}}}

\def\rmV{{\mathbf{V}}}

\def\rmX{{\mathbf{X}}}

\def\vgamma{{\bm{\gamma}}}

\def\vg{{\bm{g}}}

\def\vk{{\bm{k}}}

\def\vu{{\bm{u}}}

\def\vx{{\bm{x}}}
\def\vy{{\bm{y}}}

\DeclareMathAlphabet{\mathsfit}{\encodingdefault}{\sfdefault}{m}{sl}
\SetMathAlphabet{\mathsfit}{bold}{\encodingdefault}{\sfdefault}{bx}{n}

\def\gD{{\mathcal{D}}}

\def\gL{{\mathcal{L}}}

\def\gN{{\mathcal{N}}}
\def\gO{{\mathcal{O}}}

\def\sR{{\mathbb{R}}}

\newcommand{\E}{\mathbb{E}}

\DeclareMathOperator*{\argmin}{arg\,min}

\usepackage{subfigure}
\usepackage{graphicx}
\usepackage{subdepth}
\usepackage{subcaption}
\usepackage[export]{adjustbox}
\usepackage{caption}

\usepackage[title]{appendix}
\usepackage{bm}
\usepackage{multicol}
\usepackage{array}
\usepackage{multirow}
\usepackage{float}
\usepackage{xr}
\usepackage{thmtools, thm-restate}
\usepackage{threeparttable}
\usepackage{caption}
\usepackage{nicefrac}
\usepackage{enumerate}
\usepackage{enumitem}
\usepackage{braket}
\usepackage{chngcntr}
\usepackage{tikz}
\usepackage{ulem}
\usepackage{color, colortbl}
\usepackage{tcolorbox}

\newtheorem{theorem}{Theorem}

\newtheorem{proposition}{Proposition}

\newtheorem{lemma}{Lemma}
\newtheorem{remark}{Remark}

\hypersetup{
    urlcolor=red
}

\makeatletter
\newcommand{\labeltext}[3][]{%
    \@bsphack%
    \csname phantomsection\endcsname% in case hyperref is used
    \def\tst{#1}%
    \def\labelmarkup{\emph}% How to markup the label itself
    \def\refmarkup{}%
    \ifx\tst\empty\def\@currentlabel{\refmarkup{#2}}{\label{#3}}%
    \else\def\@currentlabel{\refmarkup{#1}}{\label{#3}}\fi%
    \@esphack%
    \labelmarkup{#2}% visible printed text.
}
\makeatother

\makeatletter
\def\blankfootnote{\xdef\@thefnmark{}\@footnotetext}
\makeatother

\newcommand{\newcheckmark}{\raisebox{0.6ex}{\scalebox{0.7}{$\sqrt{}$}}}
\newcommand{\newcrossmark}{\scalebox{0.85}[1]{$\times$}}
\newcommand*\circled[1]{\tikz[baseline=(char.base)]{\node[shape=circle,draw,inner sep=0.2pt] (char) {#1};}}
\newcommand{\rom}[1]{\uppercase\expandafter{\romannumeral #1\relax}}
\newcommand{\ours}{{\fontfamily{ppl}\selectfont Ferret}}

\begin{document}

\twocolumn[
\icmltitle{\textbf{\ours{}}: Federated Full-Parameter Tuning at Scale for Large Language Models}

% It is OKAY to include author information, even for blind
% submissions: the style file will automatically remove it for you
% unless you've provided the [accepted] option to the icml2025
% package.

% List of affiliations: The first argument should be a (short)
% identifier you will use later to specify author affiliations
% Academic affiliations should list Department, University, City, Region, Country
% Industry affiliations should list Company, City, Region, Country

% You can specify symbols, otherwise they are numbered in order.
% Ideally, you should not use this facility. Affiliations will be numbered
% in order of appearance and this is the preferred way.
\icmlsetsymbol{equal}{*}

\begin{icmlauthorlist}
\icmlauthor{Yao Shu}{equal,hkust}
\icmlauthor{Wenyang Hu}{equal,sap,nus}
\icmlauthor{See-Kiong Ng}{nus}
\icmlauthor{Bryan Kian Hsiang Low}{nus}
\icmlauthor{Fei Yu}{gml}
\end{icmlauthorlist}

\icmlaffiliation{hkust}{Hong Kong University of Science and Technology (Guangzhou)}
\icmlaffiliation{sap}{SAP}
\icmlaffiliation{nus}{National University of Singapore }
\icmlaffiliation{gml}{Guangdong Laboratory of Artificial Intelligence and Digital Economy (SZ)}

\icmlcorrespondingauthor{Yao Shu}{yaoshu@hkust-gz.edu.cn}

% You may provide any keywords that you
% find helpful for describing your paper; these are used to populate
% the "keywords" metadata in the PDF but will not be shown in the document
\icmlkeywords{Machine Learning, ICML}

\vskip 0.3in
]

% this must go after the closing bracket ] following \twocolumn[ ...

% This command actually creates the footnote in the first column
% listing the affiliations and the copyright notice.
% The command takes one argument, which is text to display at the start of the footnote.
% The \icmlEqualContribution command is standard text for equal contribution.
% Remove it (just {}) if you do not need this facility.

%\printAffiliationsAndNotice{}  % leave blank if no need to mention equal contribution
\printAffiliationsAndNotice{\icmlEqualContribution} % otherwise use the standard text.

\begin{abstract}
Large Language Models (LLMs) have become indispensable in numerous real-world applications. However, fine-tuning these models at scale, especially in federated settings where data privacy and communication efficiency are critical, presents significant challenges. Existing approaches often resort to parameter-efficient fine-tuning (PEFT) to mitigate communication overhead, but this typically comes at the cost of model accuracy. To this end, we propose \textit{\fontfamily{ppl}\selectfont federated full-parameter tuning at scale for LLMs} (\ours{}), \textbf{the first first-order method with shared randomness} to enable scalable full-parameter tuning of LLMs across decentralized data sources while maintaining competitive model accuracy. \ours{} accomplishes this through three aspects: \textbf{(\rom{1})} it employs widely used first-order methods for efficient local updates; \textbf{(\rom{2})} it projects these updates into a low-dimensional space to considerably reduce communication overhead; and \textbf{(\rom{3})} it reconstructs local updates from this low-dimensional space with shared randomness to facilitate effective full-parameter global aggregation, ensuring fast convergence and competitive final performance. Our rigorous theoretical analyses and insights along with extensive experiments, show that \ours{} significantly enhances the scalability of existing federated full-parameter tuning approaches by achieving high computational efficiency, reduced communication overhead, and fast convergence, all while maintaining competitive model accuracy. Our implementation is available at \url{https://github.com/allen4747/Ferret}.
% This therefore makes \ours{} a highly desirable solution to deploy LLMs in large-scale federated environments.

\end{abstract}

% \section{Introduction}
% Motivation: computation + number of rounds to converge

% Bonus: ACC, privacy, adaptivity

% Benefits: 
%   1. communication efficiency
%   2. Privacy-preserving is better than the standard gradient method
%   3. the best of two worlds (ZOO + FOO)
%   4. can naturally handle heterogeneity without local updates
%   5. plugin and play
%   6. more computationally efficient
%   7. can support local updates
%   8. Support adaptive optimizers (more general)
%   % 9. Support existing heterogenous techniques

\begin{figure}[t]
\centering
\hspace{-5mm}\includegraphics[width=1.05\linewidth]{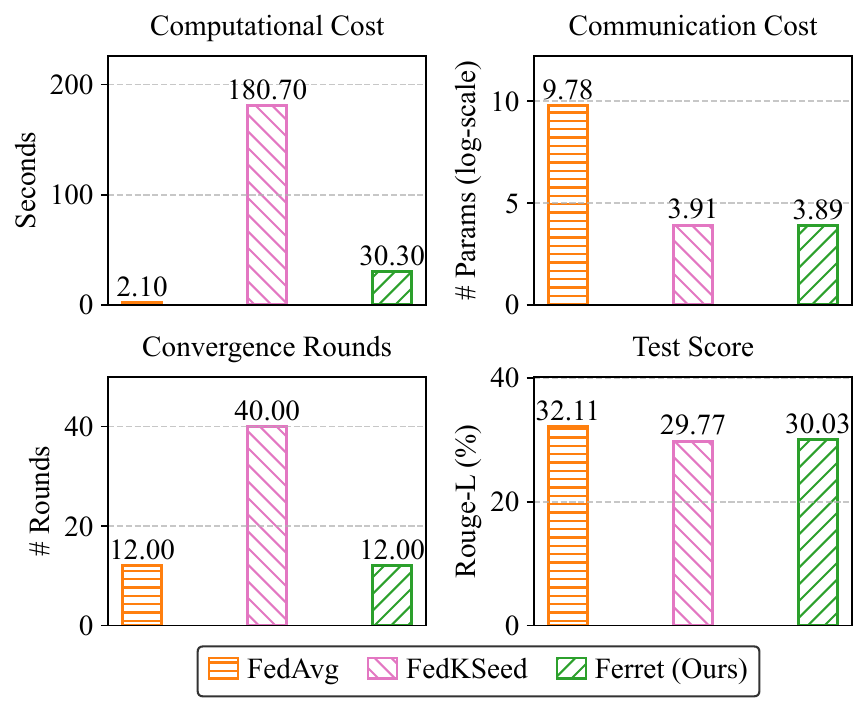}
\caption{Performance comparison of various federated full-parameter tuning algorithms on Natural Instructions dataset with LLaMA-3B. Our \ours{} shows significantly improved scalability, with a 6.0$\times$ reduction in computational cost and 3.3$\times$ fewer convergence rounds than FedKSeed, alongside a $\text{10}^\text{6}\times$ reduction in communication overhead than FedAvg, while achieving comparable test score.}
\label{fig:overview}
\vspace{-5mm}
\end{figure}

\section{Introduction}
% \blankfootnote{* Equal contribution. Correspondence to: Yao Shu <shuyao@gml.ac.cn>.}
Recently, Large Language Models (LLMs) have become indispensable tools across a wide range of real-world applications, from natural language processing tasks like translation \citep{llm-translation} and summarization \citep{llm-summarization} to more complex tasks such as code generation \citep{llm-code} and decision-making systems \citep{llm-agent}. The immense scale and versatility of LLMs make them highly valuable in practice, but they also introduce significant challenges, particularly when they are fine-tuned in federated settings. Federated Learning (FL) offers a decentralized approach to fine-tuning LLMs while retaining data on local clients to ensure privacy. However, while this approach effectively addresses privacy concerns, it also results in \textbf{prohibitive communication overhead} when the model parameters of LLMs scale to billions.

One of the straightforward strategies to mitigate the prohibitive communication costs in the federated tuning of LLMs is parameter-efficient fine-tuning (PEFT). PEFT methods \citep{lora, flexora} focus on fine-tuning only a subset of model parameters, which is able to significantly reduce the communication overhead between clients and a central server \citep{che2023federated, zhang2023fedpetuning, fedptuning, zhang2024towards}. Despite the effectiveness in reducing bandwidth usage, this type of approach often compromises \textbf{model accuracy} \citep{pu2023empirical}, as fine-tuning a subset of model parameters may fail to fully capture the nuances of local data distributions. Thus, recent efforts, e.g., FedKSeed \citep{fedkseed}, have been devoted to utilizing zeroth-order optimization (ZOO) \citep{Nesterov2017, BerahasCCS22} in federated full-parameter tuning of LLMs, aiming to maintain competitive model accuracy while reducing the communication overhead by transmitting only thousands of scalar gradients per round between clients and a central server. Unfortunately, this approach often suffers from its \textbf{poor scalability}, including \textbf{increased computational cost} per round and \textbf{a larger number of communication rounds} required for convergence, compared to FL methods that use first-order optimization (FOO), e.g., FedAvg \citep{fedavg}.

Therefore, we develop \textit{\fontfamily{ppl}\selectfont \underline{fe}de\underline{r}ated full-pa\underline{r}am\underline{e}ter \underline{t}uning at scale for LLMs} (\ours{}), \textbf{the first first-order FL approach with shared randomness} to enable scalable federated full-parameter tuning of LLMs with \textit{\fontfamily{ppl}\selectfont compelling computational efficiency}, \textit{\fontfamily{ppl}\selectfont reduced communication overhead}, and \textit{\fontfamily{ppl}\selectfont fast convergence speed}, while maintaining \textit{\fontfamily{ppl}\selectfont competitive model accuracy}, as shown in Fig.~\ref{fig:overview}. \ours{} achieves this through three aspects: First, it uses widely applied first-order methods to perform efficient local updates on each client, which usually requires fewer iterations to achieve the same local update process compared to existing ZOO-based FL. Next, \ours{} projects these updates into a low-dimensional space, resulting in a significantly reduced communication cost compared to existing FOO-based FL. Finally, \ours{} reconstructs local updates from the low-dimensional space with shared randomness for effective full-parameter global aggregation, ensuring fast convergence and competitive model accuracy compared to existing ZOO-based FL. We further complement \ours{} with rigorous theoretical analyses and insights, showing the theoretical advantages of \ours{} over other baselines and guiding the best practices for its implementation. Finally, through extensive experiments, we verify that \ours{} significantly outperforms existing methods with superior scalability and competitive model accuracy, making it a desirable solution for deploying LLMs in large-scale federated environments.

To summarize, our contributions in this work include:
\vspace{-1mm}
\begin{itemize}[topsep=0pt,leftmargin=3mm,itemsep=0pt]
    \item[$\bullet$] We novelly propose \ours{}, to the best of our knowledge, \textbf{the first first-order FL approach with shared randomness}, which \textit{significantly enhances the scalability} of federated full-parameter tuning of LLMs while maintaining \textit{competitive model accuracy}.
    \item[$\bullet$] We present \textbf{rigorous theoretical analyses and insights} to support the effectiveness of our \ours{}, demonstrating its \textit{theoretical advantages} over other baselines and \textit{guiding its best practices}.
    \item[$\bullet$] Through \textbf{extensive experiments}, we demonstrate that \ours{} consistently improves over existing methods in practice, offering both \textit{superior scalability} and \textit{competitive model accuracy}.
\end{itemize}

% Fetus represents a significant advancement in the field of federated learning, offering a practical solution for the large-scale deployment of LLMs in real-world, privacy-sensitive environments.

% \section{Related Work}

\section{Problem Setup}\label{sec:setup}
In this paper, we consider the federated full-parameter tuning of an LLM using decentralized data $\{\gD_i\}_{i=1}^{\smash{N}}$ on $N$ local clients while preserving data privacy, i.e., without sharing raw data. Specifically, given a loss function $\ell(\cdot;\cdot)$, we aim to minimize a global objective $\gL(\rvw)$ defined as the average loss across $\{\gD_i\}_{i=1}^{\smash{N}}$ over the model parameters $\rvw \in \sR^{\smash{d}}$ of an LLM. That is,
\begin{equation}
\begin{aligned}
\min_{\rvw} \gL(\rvw) 
&\triangleq \frac{1}{N} \sum_{i \in [N]} \gL^{(i)}(\rvw) \\
\text{where} \quad \gL^{(i)}(\rvw) &\triangleq \E_{\rvx^{(i)} \in \gD_i} \left[\ell(\rvw; \rvx^{(i)})\right] \ . \label{eq:problem}
\end{aligned}
\end{equation}
% \vspace{-2mm}
Following the practice in federated learning (FL), \eqref{eq:problem} can be solved through multiple rounds of local training and global aggregation. In each communication round, each client $i$ independently updates its local model parameters by minimizing its local objective $\gL^{\smash{(i)}}(\rvw)$ based on its local data $\gD_i$. After local training, the clients transmit their updated local model parameters to a central server, where they are aggregated to form an updated global model. This updated global model is then redistributed to all clients, and the process is repeated over rounds.
 
% Similar to vanilla federated learning, this goal can be achieved by several rounds of local training and global aggregation. Specifically, in each communication round, each client $i$ independently updates its local model parameters by minimizing its own local objective function $\gL^{(i)}$ on the local data $\gD_i$. After this local training, the clients send the updated local model parameters to a central server. The server then performs a global aggregation step, typically by averaging the received updates to form a new global model. This global model is then redistributed to all clients, and the process repeats for several rounds. 

% In the setting of LLM federated tuning, the key challenge of this process lies in the overheads due to the huge problem scalability we aim to solve. Specifically, the communication a

The main challenge in LLM federated full-parameter tuning is to ensure the \textbf{computational efficiency and the convergence speed} of the global model while \textbf{reducing the communication overheads}, particularly given that the parameter size $d$ of LLMs often reaches billions. While existing first-order FL \citep{fedavg, fedprox, scaffold} can ensure compelling computational efficiency and convergence speed by applying first-order updates, they typically incur $\gO(d)$ communication overheads due to the need to transmit the entire set of model parameters between clients and the central server. This type of methods hence is impractical for LLM federated full-parameter tuning due to the enormous size of LLMs. In contrast, although zeroth-order FL \citep{fedkseed} can reduce these communication costs by transmitting only several scalar gradients from their finite difference-based gradient estimation with shared randomness, they often incur more computational cost to achieve the same local update progress and a larger number of communication rounds to converge compared with first-order FL. These naturally raise the question: 
\begin{tcolorbox}[colback=orange!5!white,colframe=orange!70!white, left=0.5mm, right=1mm, top=1mm, bottom=1mm]\label{insight-1}
\textit{\fontfamily{ppl}\selectfont Can we combine the strengths of these different types of methods to achieve scalable federated full-parameter tuning of LLMs with high computational efficiency, reduced communication overhead, and fast convergence?}
\end{tcolorbox}

% The decentralized nature of the data and the heterogeneity of the local data distributions across the clients further complicate the tuning process, making it crucial to develop methods that are robust and scalable.

% \begin{figure}
% \setlength{\textfloatsep}{5pt}
\section{The \ours{} Algorithm}
To answer this question, we introduce \ours{}, \textit{\fontfamily{ppl}\selectfont \underline{fe}de\underline{r}ated full-pa\underline{r}am\underline{e}ter \underline{t}uning at scale for LLMs}, in Algo. \ref{alg:fetus}. We present an overview of \ours{} algorithm in Sec.\ref{sec:overview}, followed by a detailed explanation of its key techniques in Sec. \ref{sec:proj-update}. 

% We then evaluate the scalability and other advantages of our \ours{} in comparison to existing baselines in Sec. \ref{sec:scale}.

\subsection{Overview of \ours{}}\label{sec:overview}
To achieve scalable LLM federated full-parameter tuning, our \ours{} algorithm combines the strengths of both first-order FL, which offers efficient computation and fast convergence, and zeroth-order FL, which reduces communication overhead. Specifically, \ours{} \textit{(a)} follows first-order FL to apply first-order optimization methods for local updates on clients, ensuring both computational efficiency and fast convergence, and \textit{(b)} draws inspiration from zeroth-order FL by projecting updates into a low-dimensional space using random bases that can be regenerated using shared randomness among clients for the reconstruction of these updates, thereby reducing communication overhead. 
% This combination ensures that \ours{} achieves computational efficiency, fast convergence, and substantially reduced communication overhead in LLM federated full-parameter tuning.

Our \ours{} algorithm operates by repeating the following three sequential steps over many communication rounds, denoted by $r \in [R]$, where $R$ is the total number of rounds. For simplicity, we omit the subscript $r$ from the seeds, random bases, and projected coordinates in our notation.

\vspace{1.5mm}
\textbf{Step \circled{1}: Global Aggregation (Line 3-6 in Algo.~\ref{alg:fetus}).} 
% At the circt of the first round ($r = 1$), each client initializes its local model parameters using the pre-trained model parameters $\rvw_0$. Otherwise (for $r > 1$), each client $i \in [N]$ receives random seeds $\{s^{\smash{(i)}}\}_{i=1}^N$ and the corresponding projected coordinates $\{ \gamma_k^{\smash{(i)}} \}_{k=1}^{\smash{K}}$ for every $i \in [N]$ from the previous round, where the random seeds are then applied to generate the random bases $\{\rvv_k^{\smash{(i)}} \}_{k=1}^{\smash{K}}$ that have been employed in each client $i \in [N]$. These random bases are then applied with  the corresponding projected coordinates $\{ \gamma_k^{\smash{(i)}} \}_{k=1}^{\smash{K}}$ to reconstruct the local updates $\widetilde{\Delta}_{r-1}^{\smash{(i)}}$ from client $i \in [N]$ and the global model is updated by aggregating these local contributions as below
At the beginning of the first round ($r = 1$), each client initializes its local model parameters using the pre-trained model parameters $\rvw_0$, i.e., $\rvw_1 {\leftarrow} \rvw_0$. For subsequent rounds ($r > 1$), each client $j \in [N]$ receives the random seeds $s^{\smash{(i)}}$ and the corresponding $K$ projected coordinates $\{ \gamma_k^{\smash{(i)}} \}_{k=1}^{\smash{K}}$ of every client $i \in [N]$ from the previous round. These random seeds (i.e., shared randomness) are then used to generate $d$-dimensional random bases $\{\rvv_k^{\smash{(i)}} \}_{k=1}^{\smash{K}}$ for each client $i$. \footnote{As in \citep{fedkseed}, we can get $K$ random seeds from a single seed $s^{\smash{(i)}}$ and use them to generate $K$ random bases independently. So, one seed is sufficient for each client.} These random bases, along with the corresponding projected coordinates $\{ \gamma_k^{\smash{(i)}} \}_{k=1}^{\smash{K}}$, are applied to reconstruct local updates as $\widetilde{\Delta}_{r-1}^{\smash{(i)}}$ in every client $i$. The global model is then updated by aggregating these local contributions as follows:
\vspace{-1mm}
\begin{equation}
\begin{aligned}
\rvw_{r-1} {\leftarrow} \rvw_{r-2} {-} \frac{1}{N} \sum_{i \in [N]} \widetilde{\Delta}_{r-1}^{\smash{(i)}},
\text{with}\  \widetilde{\Delta}_{r-1}^{\smash{(i)}} {\triangleq} \sum_{k\in[K]} \gamma_k^{(i)}\rvv_k^{(i)} \ . \label{eq:global}
\end{aligned}
\end{equation}
\textbf{Step \circled{2}: Local Updates (Line 7-9 in Algo.~\ref{alg:fetus}).} After \textbf{Step \circled{1}}, each client $j$ will perform $T$-iteration first-order optimization on its local loss function by using the randomly sampled data for its local updates. Formally, if stochastic gradient descent with a local learning rate $\eta$ is used, the update rule for client $j \in [N]$ at iteration $t \in [T]$ of round $r \in [R]$ can then be represented as below with $\rvw_{r,0} \leftarrow \rvw_{r}$:
% \vspace{-1.5mm}
\begin{equation}
\rvw_{r,t}^{(j)} {\leftarrow} \rvw_{r,t-1}^{(j)} {-} \eta \nabla \ell \left(\rvw_{r,t-1}^{(j)};\rvx_{t-1}^{(j)} \right)\ . \label{eq:local}
\end{equation}
Different from the zeroth-order update in \citep{fedkseed} that requires many local update iterations,
the first-order update in \eqref{eq:local} enables each client to efficiently and effectively adapt the global model $\rvw_r$ to its specific data using a small $T$, thereby enhancing both the computational efficiency of this local update. 
% A more comprehensive elaboration will be presented in Sec.~\ref{sec:scale}.
Here, \eqref{eq:local} can be implemented using any gradient method variant, e.g., Adam \citep{kingma2014adam}.

\vspace{1.5mm}
\textbf{Step \circled{3}: Projected Updates (Line 10-12 in Algo.~\ref{alg:fetus}).} After completing the local updates above, each client $j$ randomly chooses a single new seed $s^{(j)}$ to generate $K$ new random bases $\{\rvv_k^{\smash{(j)}} \}_{k=1}^{\smash{K}}$ and employ these $K$ new random bases to project the local update $\Delta^{\smash{(j)}}_{r}$ into a $K$-dimensional coordinates $\{ \gamma_k^{\smash{(j)}} \}_{k=1}^{\smash{K}}$ based on the techniques in Sec.~\ref{sec:proj-update}.
% , aiming to considerably reduce communication overhead. 
Seed $s^{\smash{(j)}}$ and projected coordinates $\{ \gamma_k^{\smash{(j)}} \}_{k=1}^{\smash{K}}$ are then shared with other clients to facilitate the next round of global aggregation. By sharing only one single random seed and $K$ projected coordinates among $N$ clients where random bases $\{\rvv_k^{\smash{(j)}} \}_{k=1}^{\smash{K}}$ can be regenerated for global aggregation as shown in \textbf{Step \circled{1}} above, the communication overhead in LLM full-parameter tuning is therefore considerably reduced compared with first-order methods (e.g., FedAvg) especially when $T \ll d$. The communication of seed $s^{\smash{(j)}}$ can be mitigated if the same seed is used across all rounds $r\in [R]$, which can further reduce communication overhead.

% Finally, the updated parameters and computed vectors are communicated to other clients, ensuring synchronization across all participants. This process repeats until the global model converges, balancing privacy, communication efficiency, and model accuracy, making the algorithm suitable for large-scale federated learning scenarios.

\begin{figure}[t]
% \vspace{-3mm}
\begin{algorithm}[H]
\DontPrintSemicolon
\caption{\ours{}}\label{alg:fetus}
\KwIn{$\rvw_0$, $N$, $R$, $T$, $K$, $\eta$}
\For{each round $r \in [R]$}{
% \tcp*[l]{\textcolor{red}{Client-Side Update}}
% \fbox{\begin{minipage}{0.965\linewidth}
\For{each client $j \in [N]$ {\fontfamily{qpl}\selectfont\textcolor{red}{in parallel}}}{
    \If(\tcp*[h]{\fontfamily{qpl}\selectfont  \textcolor{blue}{Step \circled{1}: Global Aggregation}}){$r>1$}{
        Receive $\{s^{\smash{(i)}}\}_{i=1}^N$ and $\{\gamma_k^{\smash{(i)}}\}_{i=1,k=1}^{N,K}$\;
        
        Generate bases $\{\rvv_k^{\smash{(i)}}\}_{i=1,k=1}^{N,K}$ using $\{s^{\smash{(i)}}\}_{i=1}^N$ \;
        
        $\rvw_{r-1} {\leftarrow} \rvw_{r-2} {-} \sum_{i \in [N]}\left(\sum_{k=1}^K \gamma^{\smash{(i)}}_{k} \rvv^{\smash{(i)}}_k\right) / N$ \;
    }
    \vspace{-1mm}
    $\rvw_{r,0} \leftarrow \rvw_{r}$
    
    \For(\tcp*[h]{\fontfamily{qpl}\selectfont  \textcolor{blue}{Step \circled{2}: Local Updates}}){$t \in [T]$}{
        \vspace{1mm}
        $\rvw^{\smash{(j)}}_{r,t} \leftarrow \rvw^{\smash{(j)}}_{r,t-1} - \eta \,\nabla \ell(\rvw^{\smash{(j)}}_{r,t-1}; \rvx_{r, t-1}^{\smash{(j)}})$\;
    }

    \tcp*[l]{\fontfamily{qpl}\selectfont  \textcolor{blue}{Step \circled{3}: Projected Updates}}
    Randomly set $s^{\smash{(j)}}$ and generate bases $\{\rvv_k^{\smash{(j)}}\}_{k=1}^{\smash{K}}$

    $\Delta^{\smash{(j)}}_{r} {\leftarrow} \rvw^{\smash{(j)}}_{r-1} {-} \rvw^{\smash{(j)}}_{r}$, compute $\{\gamma_k^{\smash{(j)}}\}_{k=1}^{\smash{K}}$ with \eqref{eq:approx}

    Send $s^{\smash{(j)}}$ and $\{\gamma_k^{\smash{(j)}}\}_{k=1}^{\smash{K}}$ to the central server
}}
\end{algorithm}
\vspace{-5mm}
\end{figure}

\subsection{Update Projection and Reconstruction}\label{sec:proj-update}
% Given a random seed $s$, we can sample $n$ Gaussian random variables $\{\rvv_i\}_{i=1}^n$ (\textbf{random basis}) sequentially and independently, we then aim to approximate the ground-truth gradient $\vg$ using $\widehat{\vg} = \rmV \rvw^*$ (\textbf{inducing coordinates}):
% \begin{equation}
%     \vgamma^* = \argmin_{\vgamma} \left\|\rmV\vgamma - \Delta \rvw\right\|
% \end{equation}
% To solve it, we have $\vgamma^* = (\rmV^{\top}\rmV)^{-1} \rmV^{\top} \Delta \rvw$. May also have a random projection perspective.

% \textbf{Scalable Approximation.} Since $\rmV$ is generated from normal distribution, we can approximate the covariance matrix with $d\,\rmI$, which can be theoretically supported by our Thm.~\ref{}. We have the approximation as:
% \begin{equation}
% \begin{aligned}\label{eq:approx}
%     \widehat{\gamma}^* = \rmV^{\top} \Delta\rvw / K
% \end{aligned}
% \end{equation}

% \textbf{Seeds Allocation.}
% The Block-Wise strategy can be easily combined into other methods, such as badam. We can also employ the importance of layer to determine the number of seeds for each layer.

% As mentioned above, we will project the local updates into $K$-dimensional coordinates ($K \ll d$) to significantly reduce the communication overheads in \ours{}. To achieve this, let $\Delta$ be the local update and $\rmV = [\rvv_1\ \rvv_2\ \cdots\ \rvv_k]$ be the random bases generated by a random seed $s$, we typically aim to solve the following convex minimization problem to obtain the $K$-dimensional coordinates $\vgamma = [\gamma_1\ \gamma_2\ \cdots\ \gamma_K]^{\top}$:
As mentioned before, we aim to project the local updates into $K$-dimensional coordinates ($K \ll d$) to substantially reduce the communication overhead in LLM full-parameter tuning. To accomplish this, let $\Delta \in \sR^d$ denote any local update, and let $\rmV = [\rvv_1\ \rvv_2\ \cdots\ \rvv_K] \in \sR^{d \times K}$ represent the $K$ random bases generated by any random seed $s$, we solve the following convex minimization problem to determine the $K$-dimensional projected coordinates $\vgamma = [\gamma_1\ \gamma_2\ \cdots\ \gamma_K]^{\top}$:
\begin{equation}
\vgamma \triangleq \argmin_{\rvy} \left\|\rmV\rvy - \Delta\right\| \ . \label{eq:proj-obj}
\end{equation}
%
% Intuitively, \eqref{eq:proj-obj} seeks to minimize the difference between the projected local update $\rmV\vgamma$ and the true local update $\Delta$. 
As $\rmV$ is singular with $K \ll d$, the close-form of $\vgamma$ and its corresponding reconstruction $\widetilde{\Delta}$ will be
\begin{equation}
\vgamma = (\rmV^{\top}\rmV)^{-1}\rmV^{\top} \Delta, \quad \widetilde{\Delta} = \rmV(\rmV^{\top}\rmV)^{-1}\rmV^{\top} \Delta \ . \label{eq:close-form}
\end{equation}

% Particularly, if $\rmV$ is a rectangular matrix with ones on its main diagonal, meaning that each $\rvv_k$ is a standard basis vector, the computation of $\vgamma$ simplifies to $\vgamma = \rmV^{\top} \Delta$, which in fact corresponds to a block-wise coordinate selection for local update approximation and global aggregation. However, this method will reduce the number of parameters being updated per round, which therefore may hinder the final tuning performance. Alternatively, we can sample each element in $\rmV$ from a normal distribution $\gN(0,1)$ independently to involve all the model parameters for the global aggregation.

\textbf{Choice of Random Bases $\rmV$.} Particularly, if $\rmV$ is a rectangular matrix with ones on its main diagonal, meaning that each $\rvv_k$ is a standard basis vector, \eqref{eq:close-form} simplifies to $\vgamma = \rmV^{\top} \Delta$, which then corresponds to a block-wise dimension selection for local update projection and reconstruction. However, this approach significantly reduces the number of parameters updated per round as $K \ll d$, potentially hindering the overall tuning performance. We thus propose to sample each element in $\rvv_k \ (k\in[K])$ independently from a normal distribution with bounded 2-norm, i.e., $\left\|\rvv_k\right\| \leq 1$, aiming to realize and stabilize full-parameter tuning of LLMs for competitive overall performance. To achieve this, we can 
% \labeltext{\normalfont \textbf{(C1)}}{c1}
sample from a truncated normal distribution: $\rv \sim \gN(0, 1)$ with $\rv \in [-1/\sqrt{d},1/\sqrt{d}]$ instead.
% or \labeltext{\normalfont \textbf{(C2)}}{c2} normalize $\rvv_k$ by $\left\|\rvv_k\right\|$. 
The efficacy of this bounded norm will be demonstrated in Sec.~\ref{sec:thm-reconstruct} shortly.
% aiming to involve all model parameters in the global aggregation and thus maintain a more comprehensive global update for competitive final tuning performance. 
% The efficacy of this sampling will be demonstrated in Sec.~\ref{sec:thm-reconstruct}.

% which however makes it prohibitively costly to evaluate $\vgamma$ due to its $\gO(K^2 d + K^3)$ computational complexity.

% However, in practice, we usually randomly sample each element in $\rmV$ from a normal distribution $\gN(0,1)$, making it hard to compute $\vgamma = (\rmV^{\top}\rmV)^{-1}\rmV^{\top} \Delta$.

\textbf{Reconstruction w/o Inversion.} 
% Unfortunately, if each element in $\rmV$ is independently sampled from a normal distribution $\gN(0,1)$, \eqref{eq:close-form} involves a computational complexity of $\gO(K^2 d + K^3)$ and a memory complexity of $\gO(Kd)$, which is prohibitively costly particularly when $d$ reaches billions.
Unfortunately, 
% if each element in $\rmV$ is independently sampled from a normal distribution, 
\eqref{eq:close-form} incurs a computational complexity of $\gO(K^2 d + K^3)$ and 
storage complexity of $\gO(Kd)$ owing to the inversion of $\rmV^{\top}\rmV$ in \eqref{eq:close-form}, which is prohibitively costly, especially when $K$ is large and $d$ reaches billions. Since $\rmV^{\top}\rmV$ is a scaled empirical covariance for the aforementioned distribution of an identity covariance matrix \citep{vershynin2012close}
% : $\rvv \sim \gN(\vzero, \rmI_K)$ with $\rvv \in [-1/\sqrt{d},1/\sqrt{d}]^K$
, we propose to approximate $\rmV^{\top}\rmV$ with $\rmI_{K}$ (i.e., $K \times K$-dimensional identity matrix) and \eqref{eq:close-form} as
\begin{equation}
\vgamma \approx (\rho K)^{-1}\rmV^{\top} \Delta \ .\label{eq:approx}
\end{equation}
% Particularly, if $\rmV$ comes from \ref{c1}, 
Here, $\rho \triangleq 1 - \frac{2\psi(1/\sqrt{d})/\sqrt{d}}{2 \Phi(1/\sqrt{d}) - 1}$, where $\psi(\frac{1}{\sqrt{d}})$ and $\Phi(\frac{1}{\sqrt{d}})$ is the probability density function (PDF) and cumulative distribution function (CDF) of the standard normal distribution evaluated at $1/\sqrt{d}$, respectively. 
% If $\rmV$ comes from \ref{c2}, $\rho \triangleq 1/(d+2)$.
% Interestingly, \eqref{eq:approx} shares a similar form (i.e., $\rmV^{\top} \Delta$) to the computation of $\vgamma$ when $\rmV$ is a rectangular matrix with ones on its main diagonal. 
This approximation leads to improved computational complexity of $\gO(Kd)$ and storage complexity of $\gO(\max\{K, d\})$, where the storage complexity is reduced due to the in-place operations on random bases $\{\rvv_k\}_{k=1}^{\smash{K}}$ when computing $\{\gamma_k\}_{k=1}^{\smash{K}}$ sequentially. Consequently, we can reconstruct the true update $\Delta$ approximately using $\widetilde{\Delta}$ below
\begin{equation}
    \widetilde{\Delta} = (\rho K)^{-1}\rmV\rmV^{\top} \Delta \ , \label{eq:reconstruct}
\end{equation}
whose efficacy will be theoretically justified in Sec.~\ref{sec:thm-reconstruct}. Finally, our \eqref{eq:approx} and \eqref{eq:reconstruct} simplify the update projection and reconstruction in \eqref{eq:close-form} into straightforward matrix multiplications.

\textbf{Block-Wise Reconstruction.} The computational complexity of $\gO(Kd)$ and storage complexity of $\gO(\max\{K, d\})$ for our reconstruction in \eqref{eq:reconstruct} is still prohibitively costly, particularly for LLMs with billions of parameters. To address this, we propose a block-wise reconstruction technique to reduce both computational and storage complexities. Specifically, suppose the full dimension $d$ is divided into $L$ blocks, each with dimension $d_l$ such that $\sum_{l\in [L]} d_l =d$. Let $\Delta_l$ be the update for block $l$ and $K_l$ (with $\sum_{l\in [L]} K_l =K$) be the number of random bases allocated to this block. We propose to compute $\vgamma_l$ and  reconstruct $\Delta_l$ using random bases 
% $\rmV_l \in [-1/\sqrt{d_l},1/\sqrt{d_l}]^{d_l \times K_l}$ 
$\rmV_l$ of dimension $d_l \times K_l$ as follows:
\begin{equation}
\begin{aligned}
    \vgamma_l = (\rho_l K)^{-1}\rmV_l^{\top} \Delta_l, \quad \widetilde{\Delta}_l = (\rho_l K_l)^{-1}\rmV_l\rmV_l^{\top}\Delta_l \ . \label{eq:block-reconstruct}
\end{aligned}
\end{equation}
% Particularly, if $\rmV_l$ comes from \ref{c1}, 
Here, $\rho_l \triangleq 1 - \frac{2\psi(1/\sqrt{d_l})/\sqrt{d_l}}{2 \Phi(1/\sqrt{d_l}) - 1}$. 
% If $\rmV_l$ comes from \ref{c2}, $\rho_l \triangleq 1/(d_l+2)$. 
This trick reduces the storage complexity to $\gO(\max\{\{K_l, d_l\}_{l=1}^{\smash{L}}\})$ that is straightforward to verify, and lowers the computational complexity to $\gO(\sum_{l\in [L]}K_l d_l)$.  
Of note, \eqref{eq:block-reconstruct} also significantly reduces the computational complexity of global aggregation compared to existing methods \citep{fedkseed} (verified in Sec.~\ref{sec:exp-results}).
This block-wise reconstruction thus further enhances the scalability of our \ours{} in the federated full-parameter tuning of LLMs. Interestingly, this block-wise strategy has also been widely applied in other fields due to it efficacy \citep{sparsegpt, badam}.

% The allocation of random seeds across different layers can be systematically optimized by considering the importance of each layer to the overall model performance. This Block-Wise strategy is flexible and can be effectively combined with other optimization methods, such as Badam. By prioritizing layers of higher importance with a greater number of seeds, we enhance the accuracy of the gradient approximation while ensuring that the computational resources are utilized efficiently. This strategic allocation supports the scalability of the proposed method, allowing it to adapt to diverse model architectures and data distributions, which is crucial in heterogeneous federated learning environments.

\section{Theoretical Analyses and Insights}
% In this section, we provide rigorous theoretical analyses to substantiate the effectiveness of our \ours{}. The analyses are organized into three key components: \textit{(a)} reconstruction analysis in Sec.~\ref{sec:thm-reconstruct}, where we assess the efficacy of our reconstruction in \eqref{eq:reconstruct} under various conditions; \textit{(b)} communication round complexity in Sec.~\ref{sec:thm-round}, where we derive an upper bound on the number of communication rounds required in \ours{} for convergence; and \textit{(c)} scalability and beyond in Sec.~\ref{sec:scale}, where we highlight the scalability and other critical advantages of \ours{}.

We now provide theoretical analyses to substantiate the effectiveness of \ours{}: \textit{(a)} reconstruction analysis in Sec.~\ref{sec:thm-reconstruct}; \textit{(b)} convergence analysis in Sec.~\ref{sec:thm-round}; and \textit{(c)} scalability and beyond in Sec.~\ref{sec:scale}.

\subsection{Reconstruction Analysis}\label{sec:thm-reconstruct}
\begin{tcolorbox}[colback=blue!5!white,colframe=blue!70!white, left=0.5mm, right=1mm, top=1mm, bottom=1mm]
\begin{theorem}[\textbf{Unbiased Reconstruction}]\label{thm:unbiased}
\textit{\fontfamily{ppl}\selectfont
Given the reconstruction in \eqref{eq:reconstruct}, we have
\begin{equation*}
    \E\left[\widetilde{\Delta}\right] = \Delta \ .
\end{equation*}}
\end{theorem}
\end{tcolorbox}
% \vspace{-1mm}

To begin with, we demonstrate in Thm.~\ref{thm:unbiased} that our reconstruction in \eqref{eq:reconstruct} is unbiased, with the proof provided in Appx.~\ref{appx:proof-unbiased}. Of note, Thm.~\ref{thm:unbiased} shows that \textit{(a)} the scalar $1/(\rho K)$ is crucial for \eqref{eq:reconstruct} to achieve an unbiased reconstruction of the ground-truth update $\Delta$, and \textit{(b)} our \eqref{eq:reconstruct} avoids the bias commonly found in zeroth-order FL methods \citep{BerahasCCS22}, including FedZO \citep{fedzo} and FedKSeed \citep{fedkseed}. As a result, \eqref{eq:reconstruct} is expected to provide a more accurate update reconstruction as shown below.

\begin{tcolorbox}[colback=blue!5!white,colframe=blue!70!white, left=0.5mm, right=1mm, top=1mm, bottom=1mm]
\begin{theorem}[\textbf{Reconstruction Error}]\label{thm:reconstruct-error}
\textit{\fontfamily{ppl}\selectfont
Given the reconstruction in \eqref{eq:reconstruct}, we have
\begin{equation*}
\E\left[\left\|\widetilde{\Delta} - \Delta\right\|\right] \leq \max\left\{2\sqrt{\frac{2\ln(2d)}{\rho K}}, \frac{2\ln(2d)}{\rho K}\right\} \left\|\Delta\right\| \ .
\end{equation*}}
\end{theorem}
\end{tcolorbox}
% \vspace{-2mm}
We then demonstrate the efficacy of our reconstruction in \eqref{eq:reconstruct} by theoretically bounding the difference between the reconstructed update $\widetilde{\Delta}$ and the ground truth $\Delta$ in Thm. \ref{thm:reconstruct-error}. The proof is in Appx. \ref{appx:proof-reconstruct-error}. Of note, $1/\rho$  typically has an asymptotic rate of $\gO(d)$, which we will verify empirically in Appx.~\ref{appx:ablation-reconstruct}. 
Thm. \ref{thm:reconstruct-error} offers three critical insights of our \ours{}: \textit{(a)} Our reconstruction in \eqref{eq:reconstruct} incurs a reconstruction error at a rate of $\widetilde{\gO}(d/K)$ for $T$ local update iterations when $\sqrt{d} \geq K$, which generally aligns with the results in \citep{vershynin2010introduction}. This indicates that the reconstruction error of our \eqref{eq:reconstruct} can be linearly reduced by increasing $K$. \textit{(b)} \ours{} avoids additional constant error items \citep{BerahasCCS22} that are caused by the biased estimation in these zeroth-order FL methods, implying that our \eqref{eq:reconstruct} can be more accurate. We will justify this further in our Thm.~\ref{thm:norm} below. \textit{(c)} Thanks to the independence from the iterations (i.e., $T$) of local updates in Thm. \ref{thm:reconstruct-error}, \ours{} prevents the error accumulation over the local update iterations $T$, which is a common issue in zeroth-order FL methods \citep{fedzo, fedkseed}. 
% We will further support these advantages using the results below.

\begin{tcolorbox}[colback=blue!5!white,colframe=blue!70!white, left=0.5mm, right=1mm, top=1mm, bottom=1mm]
\begin{theorem}[\textbf{Connection with Zeroth-Order Method}]\label{thm:norm}
\textit{\fontfamily{ppl}\selectfont
Define 
$
g_k \triangleq \frac{\ell(\rvw + \epsilon \rvv_k; \rvx^{(i)}) - \ell(\rvw; \rvx^{(i)})}{\epsilon}
$, in which each element $\rv$ in $\rvv_k$ is sampled from $\rv \sim \gN(0, 1)$ with $\rv \in [-1/\sqrt{d},1/\sqrt{d}]$, $\vg \triangleq [g_1\ \cdots\ g_K]^{\top}$, and $\rmV \triangleq [\rvv_1\ \rvv_2\ \cdots\ \rvv_K] \in \sR^{d \times K}$, assume  $\ell(\cdot;\cdot)$ is $\beta$-smooth w.r.t its first argument, the zeroth-order reconstruction $\rmV\vg / K$ used in \citep{fedzo, fedkseed} then incurs:
\begin{equation*}
\left\|\frac{1}{K}\rmV\vg - \frac{1}{K}\rmV\rmV^{\top} \nabla \ell(\rvw; \rvx^{(i)})\right\| \leq \frac{1}{2}\beta \epsilon\ . 
\end{equation*}}
\end{theorem}
\end{tcolorbox}
We then show in Thm.~\ref{thm:norm} the connection between our update projection \eqref{eq:approx} and zeroth-order method used in \citep{fedzo, fedkseed}. The proof is provided in Appx.~\ref{appx:proof-norm}. Thm.~\ref{thm:norm} delivers three essential insights: \textit{(a)} When $\eps \rightarrow 0$, the reconstruction $\rmV\vg / K$ in zeroth-order method is equivalent to $\rmV\rmV^{\top} \nabla \ell(\rvw; \rvx^{(i)})/K$ and shares a similar form of \eqref{eq:reconstruct} when $\Delta$ is replaced by $\nabla \ell(\rvw; \rvx^{\smash{(i)}})$, implying that zeroth-order method in fact aims to approximate our reconstruction \eqref{eq:reconstruct}. \textit{(b)} In practice, $\epsilon>0$. So, zeroth-order method leads to a biased reconstruction with an additional error term of $\beta \eps / 2$ compared to our \eqref{eq:reconstruct}, and this error will accumulate over $T$ local iterations, implying that our \eqref{eq:reconstruct} can indeed be more accurate as we have demonstrated above. \textit{(c)} In addition, zeroth-order method is typically coupled with a single gradient (i.e., $\nabla \ell(\rvw; \rvx^{\smash{(i)}})$), whereas our \eqref{eq:reconstruct} can be applied to any vector, making it more general. Overall, these results further verify the advantages of our \eqref{eq:reconstruct} over the zeroth-order method used in \citep{fedzo, fedkseed}, which we will also support empirically in Appx.~\ref{appx:ablation-reconstruct}.

\begin{tcolorbox}[colback=blue!5!white,colframe=blue!70!white, left=0.5mm, right=1mm, top=1mm, bottom=1mm]
\begin{proposition}[\textbf{Block-Wise Reconstruction Speedup}]\label{thm:block-wise-speedup}
\textit{\fontfamily{ppl}\selectfont
For block-wise reconstruction \eqref{eq:block-reconstruct} of size $L$,
\begin{equation*}
\begin{aligned}
\sum_{l\in [L]} d_l K_l < \Bigg(\sum_{l\in [L]} d_l\Bigg) \Bigg(\sum_{l\in [L]} K_l\Bigg) = dK \ .
\end{aligned}
\end{equation*}}
\end{proposition}
\end{tcolorbox}
% \vspace{-2mm}
We next highlight the computational advantage of our block-wise reconstruction \eqref{eq:block-reconstruct} in Prop.~\ref{thm:block-wise-speedup}. The proof is in Appx.~\ref{appx:proof-block-wise-speedup}. Prop.~\ref{thm:block-wise-speedup} indicates that by dividing the reconstruction of $d$-dimensional updates into smaller blocks $\{d_l\}_{l=1}^L$, we get a reduction in overall computational complexity that is strictly less than that of the full dimension $d$ in \eqref{eq:reconstruct}. E.g., when $d_1 = \cdots = d_L$ and $K_1 = \cdots = K_L$, we have $\sum_{l\in [L]} K_l d_l = Kd/L$, showing that our block-wise reconstruction \eqref{eq:block-reconstruct} reduces the computational complexity of \eqref{eq:reconstruct} by a factor of $1/L$. This implies that increasing the number of blocks $L$ can further enhance the computational efficiency of our block-wise reconstruction \eqref{eq:block-reconstruct}. 

\begin{tcolorbox}[colback=blue!5!white,colframe=blue!70!white, left=0.5mm, right=1mm, top=1mm, bottom=1mm]
\begin{proposition}[\textbf{Block-Wise Reconstruction Error}]\label{thm:block-wise-error}
\textit{\fontfamily{ppl}\selectfont
For block-wise reconstruction \eqref{eq:block-reconstruct} of size $L$, when $\sqrt{d_l} \geq K_l$ for any $l \in [L]$,
\begin{equation*}
\begin{aligned}
\E\left[\left\|\widetilde{\Delta} - \Delta\right\|\right] < \widetilde{\gO}\Bigg(\sum_{l\in [L]} \frac{\left\|\Delta_l\right\|}{\rho_l K_l}\Bigg) \ ,
\end{aligned}
\end{equation*}
which is minimized by choosing
$
K_l \propto \sqrt{\left\|\Delta_l\right\| / \rho_l}$.}
\end{proposition}
\end{tcolorbox}
% \vspace{-2mm}
We conclude by analyzing the error induced by our block-wise reconstruction \eqref{eq:block-reconstruct} and the corresponding optimal random bases allocation in Prop.~\ref{thm:block-wise-error}. The proof is provided in Appx.~\ref{appx:proof-block-wise-error}. Prop.~\ref{thm:block-wise-error} demonstrates that reconstruction error can be minimized by adaptively allocating the number of random bases according to the gradient norm of each block. This is intuitively reasonable because a larger gradient norm typically indicates a need for more immediate model updates in practice. Hence, this insight not only provides a theoretical foundation for optimizing \ours{} but also offers practical guidance. That is, by aligning the number of random bases with gradient norms, practitioners can enhance reconstruction accuracy and overall model performance. This adaptive approach ensures efficient use of computational resources, making \ours{} versatile and effective across different datasets and federated learning scenarios.

\subsection{Convergence Analysis}\label{sec:thm-round}
In this subsection, we present the convergence of \ours{} in our Thm.~\ref{thm:convergence} below when using stochastic gradient descent (SGD) for the local updates in \eqref{eq:local}. To simplify the analysis, we primarily focus on deriving theoretical results for a homogeneous setting, where $\gL^{\smash{(i)}}(\rvw) = \gL(\rvw)$ in \eqref{eq:problem}. Results in the heterogeneous setting can be derived by following the same proof idea.

\begin{tcolorbox}[colback=blue!5!white,colframe=blue!70!white, left=0.5mm, right=1mm, top=1mm, bottom=1mm]
\begin{theorem}[\textbf{Convergence}] \label{thm:convergence}
\textit{\fontfamily{ppl}\selectfont
Define $D \triangleq \gL(\rvw_0) - \min_{\rvw}\gL(\rvw)$. Assume that $\gL(\rvw)$ is $\beta$-smooth and non-convex, and $\E[\|\nabla \gL^{\smash{(i)}}(\rvw) - \nabla\ell(\rvw;\rvx)\|^2] \leq \sigma^2$ for any $\rvx, \rvw$, when choosing $\eta \leq \frac{1}{20\beta T}$ in Algo.~\ref{alg:fetus}, the following holds for federated full-parameter tuning with $\gL^{\smash{(i)}}(\rvw) = \gL(\rvw)$,
\begin{equation*}
    \min_{r \in [R)}\E\left[\left\|\nabla \gL(\rvw_r)\right\|^2\right] \leq \gO\left(\frac{D}{\eta TR} + \eta T\sigma^2\right) \ 
\end{equation*}
where $[R)$ is the half-open interval $[0, R)$. Especially, by choosing $\eta = \frac{1}{20\beta T\sqrt{R}}$ in Algo.~\ref{alg:fetus}, the number of communication rounds are required to be $R=\gO(1/\eps^2)$ to achieve an $\eps$ convergence error.}
\end{theorem}
\end{tcolorbox}
Its proof is in Appx.~\ref{appx:proof-convergence}. Particularly, when $T=1$, Thm.~\ref{thm:convergence} recovers the result of standard SGD \citep{rsg}. Thm.~\ref{thm:convergence} provides three essential insights: \textit{(a)} Thanks to our improved update reconstruction \eqref{eq:reconstruct} as justified above, \ours{} avoids the additional constant terms accumulated over $T$ local iterations, which are typically caused by the biased gradient estimation in zeroth-order FL methods (e.g., FedZO and FedKSeed) \citep{fedzo}, thereby highlighting the superior advantage of \ours{} over these zeroth-order FL methods in convergence speed. \textit{(b)} Given a proper $\eta$, \ours{} shares the same communication round complexity as SGD, at a rate of $\gO(1/\epsilon^{\smash{2}})$, showing that the communication round complexity of \ours{} is asymptotically comparable to that of standard SGD. \textit{(c)} This communication rounds complexity is improved over that of zeroth-order FL methods \citep{fedzo} due to its independence from $d$ and other constant factors required by these zeroth-order FL methods, further highlighting the advantage of \ours{} in communication round complexity and its improved efficacy in federated full-parameter tuning over these methods.

% \subsection{Generalization Bound Via Algorithmic Stability}
\begin{table*}[t]
\caption{
Comparison of scalability (computation and communication per round, and $\#$rounds to converge) and other factors (adaptability, generalization, and privacy). Here, $d \gg K \gg T$. Symbols: $\circ$ (fewer is better), $\heartsuit$ and $\diamond$ (more is better).
}
\label{tab:comparison}
\centering
\setlength{\cmidrulewidth}{0.5pt}
% \resizebox{\textwidth}{!}{
\begin{tabular}{llllcccc}
\toprule
\multirow{2}{*}{\textbf{Method}} & \multirow{2}{*}{\textbf{Type}} & \multicolumn{3}{c}{\textbf{Scalability}} & \multicolumn{3}{c}{\textbf{Others}}\\
\cmidrule(l){3-5} \cmidrule(l){6-8} 
\noalign{\vskip 0.75mm}
& & Comp. & Comm. & $\#$Rounds  & Adapt. & Gen. & Privacy \\
\midrule
FedZO & ZOO & $\gO(\tau_0 K)$ & $\gO(d)$ & $\circ\circ\circ$ & \newcheckmark& $\heartsuit\ \heartsuit\ \heartsuit$ & $\diamond\ \diamond$ \\
\noalign{\vskip 0.75mm}
FedKSeed & ZOO & $\gO(\tau_0 K)$ & $\gO(K)$ & $\circ\circ\circ$ & \newcrossmark& $\heartsuit\ \heartsuit\ \heartsuit$ & $\diamond\diamond\diamond$ \\
\midrule
FedAvg & FOO & $\gO(\tau_1 T)$ & $\gO(d)$ & $\circ$ & \newcheckmark& $\heartsuit\ \heartsuit\ \heartsuit$ &  $\diamond$ \\
\noalign{\vskip 0.5mm}
% FedIT-SGD & $\gO(\tau_1 T)$ & $\gO(\xi d)$ & $\circ\circ\circ$ & \newcheckmark& $\circ$ & $\circ$  \\
% SCAFFOLD & & & & & & \\
% \midrule
\ours{} (ours) & FOO & $\gO(\tau_1 T)$ & $\gO(K)$ & $\circ\ \circ$ & \newcheckmark& $\heartsuit\ \heartsuit\ \heartsuit$ &  $\diamond\diamond\diamond$ \\
\bottomrule
\end{tabular}
% }
\vspace{-2mm}
\end{table*}

\subsection{Scalability and Beyond}\label{sec:scale}

With the theoretical results above, we summarize the scalability of \ours{} and compare it to existing methods like zeroth-order FL (e.g., FedZO and FedKSeed) and first-order FL (e.g., FedAvg) in Tab.~\ref{tab:comparison}.
% including computational and communication complexities per round, number of rounds to converge, and other crucial factors like adaptability, generalization, and privacy.

\textbf{Computation Per Round.} Of note, \ours{} enjoys a computational complexity of $\gO(\tau_1 T)$ for any client $i\in [N]$ per round, where $\tau_1$ is the per-iteration complexity of the first-order update (including forward and backward passes) in \eqref{eq:local}, and $T$ is the number of local iterations. This is comparable to the well-established FedAvg. In contrast, both FedZO and FedKSeed incur a complexity of $\gO(\tau_0 K)$, with $\tau_0$ being the per-iteration complexity of the zeroth-order update (i.e., forward pass) and $K$ representing the number of forward passes. As first-order updates use more accurate gradients, $T$ will be smaller than $K$ (i.e., $T \ll K$) to attain the same local update progress. Although $\tau_1$ can be at most twice $\tau_0$, our \ours{} is still more computationally efficient than FedZO and FedKSeed (see Sec.~\ref{sec:exp-results}).

\textbf{Communication Per Round.} As only one seed and $K$ projected coordinates $\{\gamma_{k}^{(i)}\}_{k=1}^{\smash{K}}$ from a client $i\in [N]$ need to be transmitted per round in Algo.~\ref{alg:fetus} with $K \ll d$, \ours{} incurs a communication overhead of $\gO(K)$, which is similar to that of FedKSeed. This is significantly more efficient than FedAvg and FedZO, which have a communication complexity of $\gO(d)$ due to their need to transmit the entire model (or gradients). This significantly reduced communication cost therefore makes \ours{} especially suitable for federated full-parameter tuning of LLMs with billions of parameters.

\textbf{Rounds to Converge.} As revealed in Sec.\ref{sec:thm-round}, our \ours{} benefits from unbiased update reconstruction in \eqref{eq:reconstruct} (validated in Thm.~\ref{thm:unbiased}), enabling fast convergence with a small number of communication rounds to achieve $\epsilon$ convergence error (see Thm.~\ref{thm:convergence}). This is significantly more efficient than zeroth-order FL methods like FedZO and FedKSeed, which require many more communication rounds to converge due to poor gradient estimation \citep{fedzo}. FedAvg, applying the ground truth local update for its global aggregation, surely converges with the fewest rounds. Overall, \ours{} remains a strong choice for federated full-parameter tuning of LLMs, even in terms of rounds to converge.

\textbf{Beyond Scalability.} Our \ours{} also offers benefits in adaptability, generalization, and privacy. Unlike FedKSeed, which is limited to SGD, \ours{} is highly adaptable, because both global aggregation \eqref{eq:global} and local update \eqref{eq:local} in \ours{} can be implemented with any gradient method variant, e.g., the widely used AdamW \citep{adamw} in LLM training. This adaptability thus makes it much easier to integrate \ours{} into existing centralized tuning workflows for LLMs, facilitating a seamless transition to federated tuning.
% Moreover, inspired by the theoretical results in \citep{train-fast}, which has shown that models trained with SGD in few iterations have minimal generalization error, our \ours{} with relatively fast convergence (refer to Thm.~\ref{thm:convergence}) 
Besides, since \ours{} enables federated tuning with full parameters, it is expected to deliver strong generalization performance as other federated full-parameter tuning methods like FedAvg, as supported in Sec.~\ref{sec:exp-results}. Finally, by transmitting only seeds and low-dimensional projected coordinates among clients, rather than the entire model (or gradients) as in FedZO and FedAvg, \ours{} ensures improved privacy for federated full-parameter tuning of LLMs.

Overall, \ours{} strikes an optimal balance between computational efficiency, communication overhead, convergence speed, and other critical factors such as adaptability, generalization, and privacy.
This makes it a scalable and desirable solution for federated full-parameter tuning of LLMs.

\section{Experiments}
% \subsection{Experimental Setup}
% \subsection{Empirical Results}
\label{sec:exp-results}

In this section, we evaluate the efficacy of \ours{}, following the practice in FedKSeed \citep{fedkseed}. We primarily compare \ours{} with other federated full-parameter tuning baselines, including both zeroth-order methods (e.g., FedZO \citep{fedzo} and FedKSeed \citep{fedkseed}) and first-order methods (e.g., FedAvg \citep{fedavg}). Our evaluations use DataJuicer-1.3B \citep{juicer} and LLaMA-3B \citep{llama} on the Natural Instructions \citep{dataset-ni} and Dolly-15K \citep{dataset-dolly} datasets, as well as larger models (i.e., LLaMA2-7B and LLaMA2-13B \citep{touvron2023llama}) on the CodeAlpaca \citep{chaudhary2023code} and GSM8K \citep{cobbe2021training}.

\textbf{FL Settings.} In each round of federated learning, 5\% of clients were randomly selected to participate. Following the same practice in FedKSeed \citep{fedkseed}, we set the total number of communication rounds to 40 for the NI dataset and 60 for Dolly-15K for all baselines. Due to the compelling efficiency of our method, we set the total number of communication rounds to 12 for the NI dataset and 20 for Dolly-15K for \ours{}. However, for more complex tasks such as CodeAlpaca and GSM8K, we run all algorithms, including our \ours{}, for 20 rounds to ensure a fair comparison. First-order baselines trained locally for one epoch, and FedKSeed trained for 200 steps, while our \ours{} algorithm trained for 10 iterations (i.e., $T=10$ in Algo.~\ref{alg:fetus}). The $K$ value was set to 4096 for FedKSeed. All approaches perform local update with a batchsize of 1 to reduce memory consumption. For each local update iteration in \ours{}, we accumulate the gradients from 4 samples.

More experimental details and ablation studies are provided in Appx.~\ref{appx:exp-setup} and Appx.~\ref{appx:ablation-reconstruct}, \ref{appx:ablation-converge} respectively.

\begin{table}[t]
\centering
\setlength{\cmidrulewidth}{0.5pt}
\caption{Comparison of Rouge-L (\%) among various algorithms. Each cell reports the mean $\pm$ std of Rouge-L scores from the final round of four runs, each with a different random seed. The highest and second-highest scores are shown in \textbf{bold} and \underline{underline}, respectively.
}
\label{tab:rouge_l_comparisons}
\resizebox{0.485\textwidth}{!}{
\begin{tabular}{lcccc}
\toprule
\multirow{2}{*}{\textbf{Algorithm}} & \multicolumn{2}{c}{\textbf{Natural Instructions}} & \multicolumn{2}{c}{\textbf{Dolly-15K}} \\
\cmidrule(l){2-3} \cmidrule(l){4-5}
& DataJuicer-1.3B & LLaMA-3B & DataJuicer-1.3B & LLaMA-3B \\ 
\midrule
FedPTuning & 19.61 $\pm$ 2.71 & 25.41 $\pm$ 1.14 & 23.98 $\pm$ 3.23 & 30.30 $\pm$ 1.16  \\
FedPrompt & \ \ 6.04 $\pm$ 0.12 & \ \ 8.95 $\pm$ 2.47 & 32.73 $\pm$ 0.87 & 24.50 $\pm$ 4.78  \\
FedIT-SGD & 19.40 $\pm$ 1.83 & 28.14 $\pm$ 0.85 & 27.23 $\pm$ 0.68 & 29.28 $\pm$ 0.50  \\
FedIT & 22.30 $\pm$ 0.42 & 28.13 $\pm$ 0.50 & 30.80 $\pm$ 0.98 & 33.23 $\pm$ 1.51  \\ 
\midrule 
FedZO & 21.74 $\pm$ 1.91 & 29.46 $\pm$ 0.38 & 26.99 $\pm$ 0.17 & 31.67 $\pm$ 0.35 \\
% FedMeZO & 21.71 $\pm$ 1.26 & 30.18 $\pm$ 0.69 & 33.32 $\pm$ 0.14 & 35.66 $\pm$ 1.06 & 33.07 $\pm$ 0.47 & 36.21 $\pm$ 0.15 \\ 
FedKSeed & 22.33 $\pm$ 1.72 & 29.77 $\pm$ 0.75 & \textbf{30.91} $\pm$ 0.29 & \underline{34.56} $\pm$ 0.28  \\
FedAvg & \underline{23.95} $\pm$ 2.76 & \textbf{32.11} $\pm$ 0.70 & 29.67 $\pm$ 1.26 & 30.98 $\pm$ 1.66 \\
% FedKSeed-Pro & 23.50 $\pm$ 1.35 & 30.19 $\pm$ 1.10 & 33.18 $\pm$ 0.68 & 36.29 $\pm$ 0.63 & 33.00 $\pm$ 0.34 & 35.95 $\pm$ 1.41 \\
\midrule
\ours{} (ours) & \textbf{24.99} $\pm$ 0.99 & \underline{30.03} $\pm$ 0.99 & \underline{30.63} $\pm$ 0.84 & \textbf{34.57} $\pm$ 0.57\\
\bottomrule
\end{tabular}
}
% \vspace{-3mm}
\end{table}
\begin{table}[t]
\centering
\setlength{\cmidrulewidth}{0.5pt}
\caption{More comparison of Rouge-L (\%) among various algorithms. Each cell reports the mean $\pm$ std of Rouge-L scores from the final round of four runs, each with a different random seed. The highest and second-highest scores are shown in \textbf{bold} and \underline{underline}, respectively.
}
\label{tab:rouge_l_comparisons_7b_13b}
\resizebox{0.485\textwidth}{!}{
\begin{tabular}{lcccc}
\toprule
\multirow{2}{*}{\textbf{Algorithm}} & \multicolumn{2}{c}{\textbf{CodeAlpaca}} & \multicolumn{2}{c}{\textbf{GSM8K}} \\
\cmidrule(l){2-3} \cmidrule(l){4-5}
& LLaMA2-7B & LLaMA2-13B & LLaMA2-7B & LLaMA2-13B \\ 
\midrule
FedIT & 4.66 $\pm$ 0.18 & 6.10 $\pm$ 0.18 & 30.31 $\pm$ 0.29 & 13.46 $\pm$ 0.34 \\
\midrule
FedZO & 4.58 $\pm$ 0.26 & 6.19 $\pm$ 0.32 & 30.41 $\pm$ 0.31 & 13.63 $\pm$ 0.34 \\
FedKSeed & 8.33 $\pm$ 0.98 & 10.70 $\pm$ 0.47 & 28.26 $\pm$ 3.60 & 33.67 $\pm$ 1.15  \\
FedAvg & \textbf{15.41} $\pm$ 0.43 & \textbf{14.68} $\pm$ 0.26 & \textbf{38.30} $\pm$ 0.40 & \textbf{39.82} $\pm$ 0.17 \\
\midrule
\ours{} (ours) & \underline{12.10} $\pm$ 0.47 & \underline{11.84} $\pm$ 0.91 & \underline{36.10} $\pm$ 1.18 & \underline{34.50} $\pm$ 1.42\\
\bottomrule
\end{tabular}
}
\vspace{-2mm}
\end{table}

\begin{table}[t]
\centering
\setlength{\cmidrulewidth}{0.5pt}
\caption{Comparison of per-round computational cost and communication overhead on LLaMA-3B, including (a) the computational costs of local updates, global aggregation, and overall cost per round; and (b) the per-round communication cost. The improvement achieved by our \ours{} is reported in brackets using \textcolor{blue}{blue} (compared with FedKSeed) and \textcolor{orange}{orange} (compared with FedAvg).}
\resizebox{0.485\textwidth}{!}{
\begin{tabular}{lcccc}
\toprule
\multirow{2}{*}{\textbf{Algorithm}}
% & \multirow{2}{*}{\textbf{Type}} 
& \multicolumn{3}{c}{\textbf{Computational Cost} (Sec.)} & \multicolumn{1}{c}{\textbf{Communication Cost}} \\
\cmidrule(l){2-4} 
 & Local Update & Global Aggr. & Overall & (\# params.) \\
\midrule
% FedAvg & 22.9 & 1.3$\times$10$^9$ & 61.7 & 3.0$\times$10$^9$ \\
\cellcolor{gray!20} FedZO  & \cellcolor{gray!20} 32.6 & \cellcolor{gray!20} 0.3 & \cellcolor{gray!20} 32.9 & \cellcolor{gray!20} 6.0$\times$10$^9$ \\
% FedMeZO & 24.5 & & & 3.0$\times$10$^9$ & \\
% \midrule
% FedPTuning & & & &  \\
% FedPrompt & & & &  \\
% FedIT-SGD & & & &  \\
% FedIT & & & &  \\
\cellcolor{blue!20} FedKSeed  &\cellcolor{blue!20} 56.9 & \cellcolor{blue!20} 123.8 &\cellcolor{blue!20} 180.7  &\cellcolor{blue!20} 8.2$\times$10$^3$  \\
% FedKSeed-Pro & & & &  \\
% FedAvg 40 step local update 2.3 global 4.8
% FedZO local: 58.1 global 4.8

 \cellcolor{orange!20} FedAvg  &  \cellcolor{orange!20} 1.8 &  \cellcolor{orange!20} 0.3 & \cellcolor{orange!20} 2.1 &  \cellcolor{orange!20} 6.0$\times$10$^9$ \\
\midrule
\ours{} (ours) & \textbf{5.6} \textcolor{blue}{(10.2$\times$)} & \textbf{24.7} \textcolor{blue}{(5.0$\times$)} & \textbf{30.3} \textcolor{blue}{(6.0$\times$)} & \textbf{7.8$\times$10$^3$} \textcolor{orange}{(10$^6\times$)} \\
\bottomrule
\end{tabular}
}
\label{tab:computation:cost}
% \vspace{-4mm}
\end{table}

\begin{table}[t]
\centering
\setlength{\cmidrulewidth}{0.5pt}
\caption{Comparison of per-round computational cost and communication overhead on LLaMA2-7B, focusing on (a) the computational costs from local updates, global aggregation, and the overall cost per round; and (b) the per-round communication cost. The improvement achieved by our \ours{} is reported in brackets using \textcolor{blue}{blue} (compared with FedKSeed) and \textcolor{orange}{orange} (compared with FedAvg).}
\resizebox{0.485\textwidth}{!}{
\begin{tabular}{lcccc}
\toprule
\multirow{2}{*}{\textbf{Algorithm}}
& \multicolumn{3}{c}{\textbf{Computational Cost} (Sec.)} & \multicolumn{1}{c}{\textbf{Communication Cost}} \\
\cmidrule(l){2-4}   
 & Local Update & Global Aggr. & Overall & (\# params.) \\
\midrule
\cellcolor{gray!20} FedZO  & \cellcolor{gray!20} 54.1 & \cellcolor{gray!20} 0.7 & \cellcolor{gray!20} 54.8 & \cellcolor{gray!20} 1.4$\times$10$^{10}$ \\
\cellcolor{blue!20} FedKSeed  &\cellcolor{blue!20} 117.0 & \cellcolor{blue!20} 510.0 &\cellcolor{blue!20} 627.0 &\cellcolor{blue!20} 8.2$\times$10$^3$  \\
 \cellcolor{orange!20} FedAvg  &  \cellcolor{orange!20} 5.8 &  \cellcolor{orange!20} 0.7 & \cellcolor{orange!20} 6.5 &  \cellcolor{orange!20} 1.4$\times$10$^{10}$ \\
\midrule
\ours{} (ours) & \textbf{8.9} \textcolor{blue}{(13.1$\times$)} & \textbf{88.3} \textcolor{blue}{(5.8$\times$)} & \textbf{97.2} \textcolor{blue}{(6.5$\times$)} & \textbf{6.4$\times$10$^3$} \textcolor{orange}{(10$^6\times$)} \\
\bottomrule
\end{tabular}
}
\label{tab:computation:cost:7b_gsm8k}
\vspace{-3mm}
\end{table}

\begin{figure}[t]
\vspace{-2mm}
\centering
\begin{tabular}{cc}
    \hspace{-6mm}
    \includegraphics[width=0.245\textwidth]{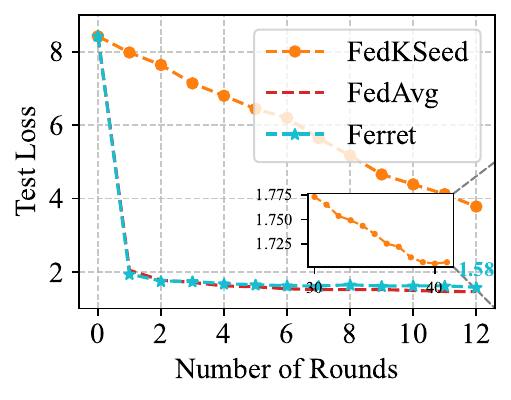}&
    \hspace{-4mm}
    \raisebox{0.0mm}{\includegraphics[width=0.25\textwidth]{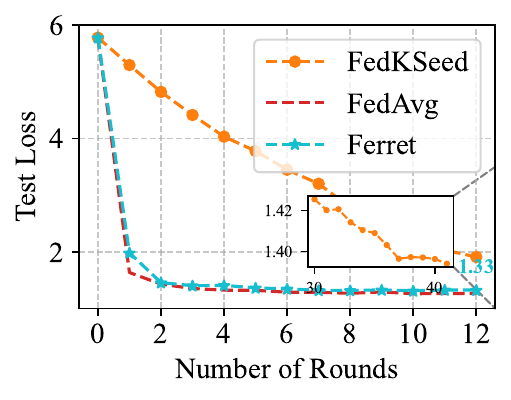}} \\
    \hspace{-4mm}
    % \includegraphics[width=0.33\textwidth]{workspace/figs/llama-7b.pdf} \\
    % \raisebox{0.2mm}{\includegraphics[natwidth=400,natheight=300]{workspace/figs/llama-3b.pdf}} \\
    {(a) DataJuicer-1.3B} & {\hspace{0mm}(b) LLaMA-3B} 
    % & {\hspace{0mm}(c) LLaMA2-7B}
\end{tabular}
\vspace{-2mm}
\caption{
Comparison of communication rounds required by \ours{}, FedKSeed, and FedAvg for convergence on Natural Instructions with (a) DataJuicer-1.3B and (b) LLaMA-3B.
}
\label{fig:rounds}
\vspace{-3mm}
\end{figure}

\subsection{Comparison on Accuracy}\label{sec:exp-accuracy}
We present the model accuracy of different federated tuning methods in Tab.~\ref{tab:rouge_l_comparisons} and \ref{tab:rouge_l_comparisons_7b_13b}. The results in Tab.~\ref{tab:rouge_l_comparisons} demonstrate that federated full-parameter tuning methods (including FedAvg, FedZO, FedKSeed, and \ours{}) generally achieve better model accuracy compared to PEFT-based federated tuning methods (such as FedPTuning, FedPrompt, FedIT-SGD, and FedIT). This underscores the importance of full-parameter tuning for Large Language Models (LLMs).
Importantly, the results in both tables show that our proposed method consistently delivers strong or competitive performance across four different scenarios. Specifically, on the Natural Instructions dataset, our method outperforms all others for different model sizes, with up to a 2.66\% improvement over the next best method, FedKSeed. On the Dolly-15K dataset, our method maintains competitive performance. Moreover, on both the CodeAlpaca and GSM8K datasets, our method achieves noticeably improved accuracy over FedIT and other zeroth-order baselines (i.e., FedZO and FedKseed). Of note, \ours{} slightly underperform FedAvg, likely due to reconstruction errors caused by our method for these complex tasks. Overall, these results have well demonstrated the ability of our method to sustain strong model accuracy in practice across various datasets and model sizes.

\subsection{Comparison on Scalability} \label{sec:exp-scalability}
Since we focus on federated full-parameter tuning of LLMs, we primarily provide a detailed scalability comparison of this type of methods, including FedZO, FedKSeed, FedAvg, and \ours{}. We evaluate their scalability performance on Natural Instructions using LLaMA-3B (see Tab.~\ref{tab:computation:cost}) and GSM8K using LLaMA2-7B (see Tab.~\ref{tab:computation:cost:7b_gsm8k}), where the calculation of computational cost and communication overhead is in Appx.~\ref{appx:comp&comm} and more comparison on LLaMA2-13B is in Appx.~\ref{appx:more-comparison}. The results in Tab.~\ref{tab:computation:cost} and Tab.~\ref{tab:computation:cost:7b_gsm8k} demonstrate that compared with FedKSeed, \ours{} achieves substantial reductions in computational costs: a 10.2$\times$ improvement for local updates on LLaMA-3B and 13.1$\times$ on LLaMA2-7B, a 5.0$\times$ improvement in global aggregation on LLaMA-3B and 5.8$\times$ on LLaMA2-7B, as well as a 6.5$\times$ improvement for overall computational cost per round on LLaMA-3B and 6.8$\times$ on LLaMA2-7B. 
% \footnote{The reduced improvement in overall tuning cost for our \ours{} on LLaMA2-7B, compared to LLaMA-3B, is because that both \ours{} and FedKSeed are using the same number of communication rounds for more complex tasks such as GSM8K.}
These advancements stem from several key innovations: our first-order local updates, which reduce the number of required iterations; block-wise reconstruction, which optimizes global aggregation; and precise reconstruction, which significantly decreases communication round complexity. Furthermore, compared to FedAvg that does not leverage any shared randomness, \ours{} exhibits an enormous reduction in overall communication costs, i.e., 10$^6\times$ on LLaMA-3B and 10$^7\times$ on LLaMA2-7B. This emphasizes the ability of \ours{} in scaling federated full-parameter tuning. 

In Fig.~\ref{fig:rounds}, we also compare the convergence speeds of our \ours{} with other baselines (e.g., FedKSeed and FedAvg) on Natural Instructions (with DataJuicer-1.3B and LLaMA-3B). The findings show that, \ours{} converges remarkably fast, requiring only two communication rounds in line with FedAvg compared to the 40 rounds needed by FedKSeed. This results in a 20$\times$ reduction in communication round complexity for both DataJuicer-1.3B and LLaMA-3B. 
% Even on larger models like LLaMA2-7B, \ours{} maintains a comparable convergence speed to FedAvg, which is still considerably faster than FedKSeed. 

% Overall, these results highlight the scalability of \ours{}, as discussed in Sec.~\ref{sec:scale}, and demonstrate its ability to balance computational efficiency, communication overhead, and fast convergence.

\subsection{Comparison on GPU Memory Consumption}

\begin{table}[t]
\centering
\caption{Peak GPU memory footprint of different methods.}
\label{tab:memory:cost:1b_3b_NI}
\begin{tabular}{l|r|r}
\toprule
\textbf{Method} & DataJuicer-1.3B & LLaMA-3B \\
\midrule
FedAvg & 9.9 GB & 19.1 GB \\
FedKSeed & 3.5 GB & 7.8 GB \\
\ours{} (ours) & 9.9 GB & 19.1 GB \\
\bottomrule
\end{tabular}
\vspace{-4mm}
\end{table}

We further provide the comparison on GPU memory consumption for different methods in Tab.~\ref{tab:memory:cost:1b_3b_NI}. The results demonstrate that our proposed method \ours{} maintains the same peak GPU memory footprint as FedAvg, requiring 9.9 GB and 19.1 GB for DataJuicer-1.3B and LLaMA-3B models, respectively, on the Natural Instructions dataset. Although FedKSeed, as a zeroth-order method, exhibits lower memory requirements, our \ours{} delivers superior converged performance (refer to Sec.~\ref{sec:exp-accuracy}) and improved scalability (refer to Sec.~\ref{sec:exp-scalability}) without introducing additional memory overhead compared to the baseline FedAvg method.

\subsection{Ablation Studies}
\begin{figure}[t]
\vspace{-2mm}
\centering
\begin{tabular}{cc}
    \hspace{-1mm}
    \includegraphics[width=0.48\columnwidth]{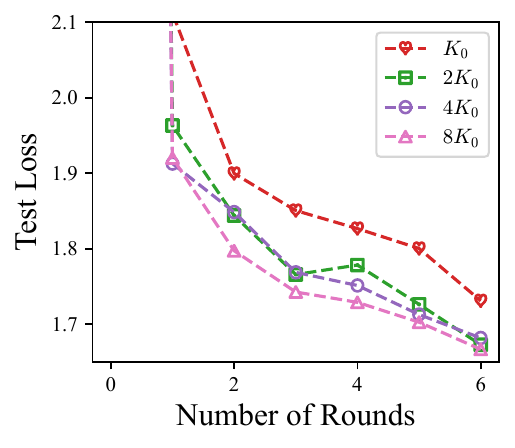}&
    \hspace{-4mm}
    \raisebox{2.4mm}{
    \includegraphics[width=0.48\columnwidth]{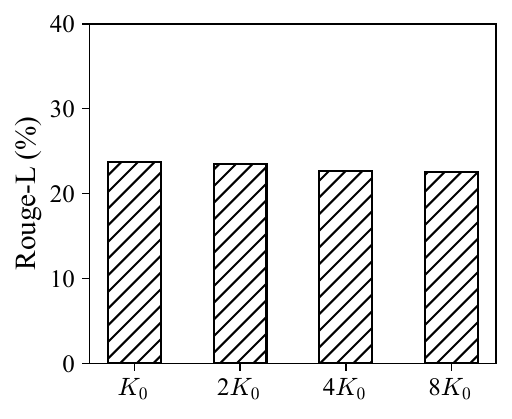}
    } 
    \\
    \vspace{-0.2mm}
    {(a) Convergence} & {(b) Generalization}
\end{tabular}
\vspace{-3mm}
\caption{
Convergence and generalization of our \ours{} under varying $K$ on Natural Instructions with DataJuicer-1.3B where $2K_0$ corresponds to the communication cost of 7.8$\times$10$^3$ per round in Tab.~\ref{tab:computation:cost}. 
}
\label{fig:converge-K}
\vspace{-3mm}
\end{figure}

\textbf{Convergence and Generalization of \ours{} under Varying $K$.} In Fig.~\ref{fig:converge-K}, we present the convergence and generalization of \ours{} under varying $K$ on the Natural Instructions dataset with DataJuicer-1.3B, using the same experimental setup as described in Appx.~\ref{appx:exp-setup}. Notably, Fig.~\ref{fig:converge-K} shows that: \textit{(a)} a larger number of random bases (i.e., a larger $K_0$) generally leads to improved convergence, while the generalization performance remains comparable; \textit{(b)} $2K_0$ already provides compelling convergence and generalization performance, and further increasing $K$ yields only marginal improvements in convergence; and \textit{(c)} a slight decrease in generalization performance as $K$ increases is likely due to the reduced regularization effect from noisy gradients.

\begin{figure}[t]
\vspace{-2mm}
\centering
\begin{tabular}{cc}
    \hspace{-4mm}
    \includegraphics[width=0.5\columnwidth]{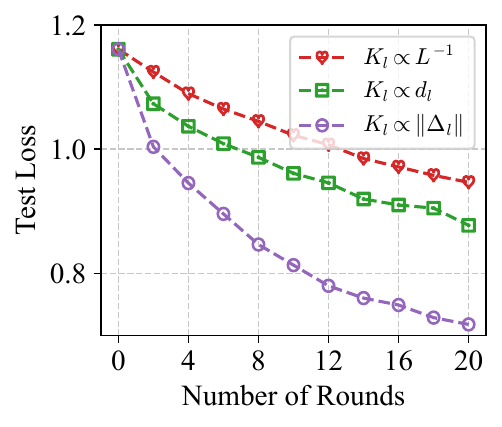}&
    \hspace{-4mm}
    \raisebox{2.2mm}{
    \includegraphics[width=0.49\columnwidth]{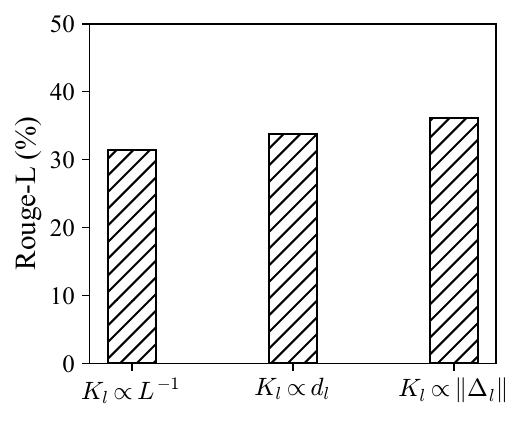}} \\
    {(a) Convergence} & {(b) Generalization}
\end{tabular}
\vspace{-2mm}
\caption{
Convergence and generalization of \ours{} under varying allocation scheme of $K$ for our block-wise reconstruction, in which $K_l \propto \left\|\Delta_l\right\|$ corresponds to our results in the previous sections. 
}
\label{fig:alloc}
\vspace{-5mm}
\end{figure}

\textbf{Convergence and Generalization of \ours{} under Varying Allocation of $K$.} In Fig.~\ref{fig:alloc}, we present the convergence and generalization of \ours{} under different allocation schemes for $K$ in our block-wise reconstruction, using the same experimental setup described in Appx.~\ref{appx:exp-setup}, where $K_l \propto \left\|\Delta_l\right\|$ corresponds to our results in Sec.~\ref{sec:exp-results}. Notably, Fig.~\ref{fig:alloc} shows that \ours{} achieves both faster convergence and improved generalization performance by following the best practices guided by Prop.~\ref{thm:block-wise-error}. These findings therefore validate the significance and correctness of our Prop.~\ref{thm:block-wise-error}.

\begin{figure}[t]
\vspace{-1mm}
\centering
\begin{tabular}{cc}
    \hspace{-4.5mm}
    \includegraphics[width=0.46\columnwidth]{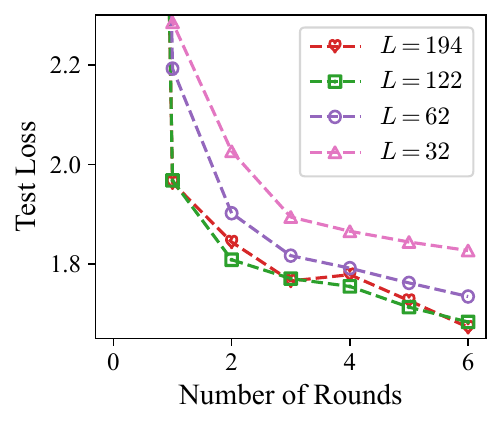}&
    \hspace{-4mm}
    \raisebox{2.5mm}{\includegraphics[width=0.525\columnwidth]{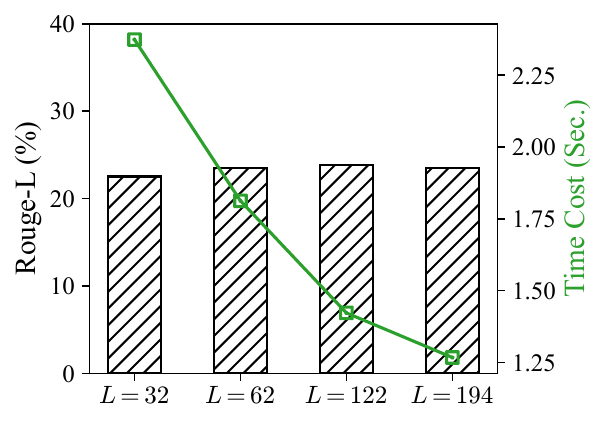}} \\
    {(a) Convergence} & {\hspace{-4mm} (b) Generalization \& Time Cost}
\end{tabular}
\vspace{-2mm}
\caption{
Convergence, generalization, and projection time cost per round of \ours{} under varying $L$ on Natural Instructions with DataJuicer-1.3B where $L=194$ is used in Tab.~\ref{tab:rouge_l_comparisons}. 
}
\label{fig:converge-L}
\vspace{-5mm}
\end{figure}

\textbf{Convergence and Generalization of \ours{} under Varying $L$.} In Fig.~\ref{fig:converge-L}, we present the convergence, generalization, and projection time cost of \ours{} under varying block sizes $L$ on the Natural Instructions dataset with DataJuicer-1.3B, using the same experimental setup as described in Appx.~\pageref{appx:exp-setup}. Notably, Fig.~\ref{fig:converge-L} shows that increasing the number of blocks (i.e., a larger $L$) leads to improved convergence and reduced time cost for projection and reconstruction, while the generalization performance remains comparable. This improved convergence is likely due to the logarithmic term in the reconstruction error of our \eqref{eq:reconstruct}, as a larger number of blocks reduces the dimensionality of each block, thereby minimizing reconstruction error. In addition, the reduced time cost aligns with our analysis in Sec.~\ref{sec:thm-reconstruct} and the empirical results shown in Fig.~\ref{fig:ablation-reconstruct}(c), further highlighting the efficacy of our block-wise reconstruction method \eqref{eq:block-reconstruct}.

\section{Conclusion}

In this paper, we introduce \ours{}, which offers a highly desirable solution for scalable federated full-parameter tuning of LLMs. By achieving high computational efficiency, fast convergence, and reduced communication overhead, \ours{} overcomes the limitations of existing methods, striking an improved balance among these critical factors. Moreover, our rigorous theoretical analyses and extensive experiments validate \ours{} as a robust and reliable approach for deploying LLMs in large-scale federated settings. While our work focuses on the performance and efficiency of \ours{}, a critical area for future research is the in-depth privacy analysis of our method and exploring privacy-preserving mechanisms to ensure data security.

\section*{Acknowledgements}
This research/project is supported by the National Research Foundation, Singapore under its National Large Language Models Funding Initiative (AISG Award No: AISG-NMLP-$2024$-$001$). Any opinions, findings and conclusions or recommendations expressed in this material are those of the author(s) and do not reflect the views of National Research Foundation, Singapore. This research/project is supported by SAP and Singapore’s Economic Development Board under the Industrial Postgraduate Programme.
% National Research Foundation Singapore and DSO National Laboratories under the AI Singapore Programme (AISG Award No: AISG$2$-RP-$2020$-$018$).  

\section*{Impact Statement}
This paper presents work whose goal is to advance the field of federated learning for large language models. There are many potential societal consequences of our work, none of which we feel must be specifically highlighted here.

% \clearpage
\bibliography{icml2025}
\bibliographystyle{icml2025}

\clearpage
\begin{appendices}
\onecolumn
\section{Related Work}\label{sec:related-works}

\textbf{Federated Parameter-Efficient Fine-Tuning (PEFT) for LLMs.}
% The field of federated learning (FL) has gained significant traction in its application to the fine-tuning of large language models (LLMs). Traditional FL approaches in this domain \citep{zhang2023fedpetuning, fedptuning, FedIT, FedPrompt} have predominantly focused on parameter-efficient fine-tuning (PEFT) techniques \citep{lora, flexora, lester2021power, instinct, zopo}, which reduce the number of trainable parameters in LLMs to mitigate the extensive communication overheads in FL scenarios. Unfortunately, while PEFT methods such as those proposed in \citep{fedptuning, FedIT} have shown promise, they often fall short in achieving the accuracy levels possible with full-parameter tuning \citep{pu2023empirical}, particularly in non-IID (non-independent and identically distributed) data settings commonly encountered in FL. In contrast, this paper focuses on federated full-parameter tuning of LLMs, aiming to achieve significantly reduced communication overhead with competitive model accuracy.
Federated learning (FL) has become a crucial approach for training large language models (LLMs) while preserving data privacy. A significant body of work in this area has focused on parameter-efficient fine-tuning (PEFT) techniques \citep{lora, flexora, lester2021power, instinct, zopo, lau2024dipper} to address the substantial communication costs associated with training massive LLMs in distributed settings. These methods, such as those explored in \citep{zhang2023fedpetuning, fedptuning, FedIT, FedPrompt}, aim to reduce the number of trainable parameters, thereby decreasing communication overhead. While PEFT approaches offer a viable solution, they often compromise on model accuracy compared to full-parameter fine-tuning, particularly in challenging non-IID (non-independent and identically distributed) data scenarios common in FL \citep{pu2023empirical}. In contrast, our work tackles the challenge of federated full-parameter tuning of LLMs, aiming to achieve accuracy levels comparable to centralized training without the prohibitive communication costs. This gap in achieving optimal accuracy with existing federated PEFT methods underscores the need for our approach. 

\textbf{Federated Learning with Gradient Compression.} Another line of research explores gradient compression techniques to reduce communication overhead in federated learning. Methods like top-k sparsification \citep{top-k}, signSGD \citep{sign-sgd}, and sketching-based compression \citep{fetchsgd} aim to reduce the size of transmitted gradients. Yet, these methods often achieve limited compression ratios and may not be effective for the high-dimensional parameter spaces of LLMs, particularly during full-parameter tuning. The millions of parameters involved in full-parameter LLM tuning far exceed the compression capabilities of these approaches. Furthermore, these techniques are not specifically tailored for the unique characteristics of LLMs, potentially limiting their effectiveness. In contrast, our proposed method leverages shared randomness within first-order FL to achieve significantly higher compression rates, scaling down the communication cost by orders of magnitude. This specialization for LLMs provides a critical advantage.

\textbf{Federated Learning with Shared Randomness.} The use of shared randomness has emerged as a promising avenue for reducing communication costs in FL. Methods like \citep{fedkseed, xu2023federated, maritan2023fedzen, feng2023does, dorfman2023docofl, zelikman2023just, rahimi2024evofed} demonstrate that transmitting only random seeds and scalar gradients can drastically reduce communication overhead. However, these methods typically rely on zeroth-order optimization (ZOO) for local updates on each client. ZOO methods, while communication-efficient, are computationally expensive, requiring more rounds to converge than first-order methods like FedAvg \citep{fedavg} and FedProx \citep{fedprox}. This computational inefficiency hinders scalability, particularly in large-scale federated environments. In contrast, our work introduces the use of shared randomness within first-order FL, combining the communication efficiency of shared randomness with the computational efficiency of first-order optimization. To the best of our knowledge, this is the first time that shared randomness has been successfully integrated with first-order FL for LLM fine-tuning, addressing a critical gap in the field.

% Overall, in contrast to these existing methods, Fetus introduces a novel approach that combines the benefits of full-parameter tuning with advanced communication strategies specifically tailored for the unique demands of LLMs in federated settings. By employing adaptive communication protocols, model compression techniques, and efficient gradient transfer methods, Fetus addresses the critical challenges of communication overhead and computational efficiency without compromising model performance. This positions Fetus as a scalable and practical solution for federated LLM tuning, particularly in privacy-sensitive and bandwidth-limited environments.

\section{Proofs}\label{appx:proofs}
\subsection{Proof of Thm.~\ref{thm:unbiased}}\label{appx:proof-unbiased}

Suppose $\rv$ is randomly and independently sampled from a truncated normal distribution, i.e.,  $\rv \sim \gN(0, 1)$ with $\rv \in [-1/\sqrt{d},1/\sqrt{d}]$, we have
\begin{equation}
\begin{aligned}
\E\left[\rv\right] &= 0 \ ,
\end{aligned}
\end{equation}
and also
\begin{equation}
\begin{aligned}
\E\left[\rv^2\right] &= \left(\E\left[\rv\right]\right)^2 + \text{VAR}(\rv) \\[7pt]
&= \text{VAR}(\rv) \\[1pt]
&= 1 - \frac{1/\sqrt{d}(\psi(1/\sqrt{d}) + \psi(-1/\sqrt{d}))}{\Phi(1/\sqrt{d}) - \Phi(-1/\sqrt{d})} - \left(\frac{\psi(1/\sqrt{d}) - \psi(-1/\sqrt{d})}{\Phi(1/\sqrt{d}) - \Phi(-1/\sqrt{d})}\right)^2 \\
&= 1 - \frac{2\psi(1/\sqrt{d})/\sqrt{d}}{2\Phi(1/\sqrt{d}) - 1}
\end{aligned}
\end{equation}
where $\psi(\frac{1}{\sqrt{d}})$ and $\Phi(\frac{1}{\sqrt{d}})$ is the probability density function (PDF) and cumulative distribution function (CDF) of the standard normal distribution evaluated at $1/\sqrt{d}$, respectively. 

According to Sec.~\ref{sec:proj-update}, each element $\rv$ in $\rmV$ is randomly and independently sampled from the truncated normal distribution above. We therefore have the following to conclude our proof:
\begin{equation}
% \begin{aligned}
\E\left[\widetilde{\Delta}\right] = \frac{1}{\rho K}\E\left[\rmV\rmV^{\top}\right] \Delta =  \frac{1}{\rho K}\E\left[\sum_{k=1}^K \rvv_k \rvv_k^{\top}\right] \Delta = \Delta \ .
% &= \Delta \ ,
% \end{aligned}
\end{equation}

% For \ref{c2}, as each element $\rv$ in $\rvv_k$ is first independently sampled from a normal distribution and then scaled by $\left\|\rvv_k\right\|$, $\rvv_k$ can then be taken as a sample from a uniform distribution on the unit sphere. So, we have
% \begin{equation}
% \begin{aligned}
% \pi &= \text{VAR}(\rv) \\
% &= \frac{1}{d+2} \\[3pt]
% &= \rho \ ,
% \end{aligned}
% \end{equation}

\subsection{Proof of Thm.~\ref{thm:reconstruct-error}}\label{appx:proof-reconstruct-error}
To begin with, we introduce the lemma below to ease our proof.
\begin{lemma}[Matrix Bernstein Inequality, Thm. 1.6.2 in \citep{matrix-concent}]\label{le:Bernstein}
Let $\rmX_1,\cdots,\rmX_K$ be independent, zero mean, and symmetry matrices of size $d \times d$, if $\left\|\rmX_k\right\| \leq C$ for any $k \in [K]$, we then have
\begin{equation}
    \E\left[\left\|\sum_{k=1}^K \rmX_k \right\|\right] \leq \sqrt{2\nu\ln(2d)} + \frac{1}{3}C\ln(2d)
\end{equation}
where $\nu \triangleq \left\|\sum_{k=1}^K \E\left[\rmX_k^2\right]\right\|$.
\end{lemma}

Define $\rmX_k \triangleq \left(\rvv_k\rvv_k^{\top} - \rho\rmI_d\right) / K$, We have
\begin{equation}
\begin{aligned}
    \left\|\rmX_k\right\| &\stackrel{(a)}{=} \frac{1}{K}\left\|\rvv_k\rvv_k^{\top} - \rho \rmI_d\right\| \\
    &\stackrel{(b)}{\leq} \frac{1}{K}\left(\left\|\rvv_k\rvv_k^{\top}\right\| + \rho \left\|\rmI_d\right\|\right) \\
    &\stackrel{(c)}{=} \frac{1}{K}\left(\left\|\rvv_k^{\top}\rvv_k\right\| + \rho\right) \\
    &\stackrel{(d)}{\leq} \frac{2}{K}
\end{aligned}
\end{equation}
where $(b)$ comes from triangle inequality and $(c)$ is due to the fact that outer product $\rvv_k\rvv_k^{\top}$ and inner product $\rvv_k^{\top}\rvv_k$ shares the same operator norm. Finally, $(d)$ results from $\rho < 1$ and $\rvv_k^{\top}\rvv_k \leq 1$.

Besides, we also have
\begin{equation}
\begin{aligned}
    \E\left[\rmX_k^2\right] &\stackrel{(a)}{=} \frac{1}{K^2}\E\left[\rvv_k\rvv_k^{\top}\rvv_k\rvv_k^{\top} - 2\rho\rvv_k\rvv_k^{\top} + \rho^2 \rmI_d\right] \\[3pt]
    &\stackrel{(b)}{\preceq} \frac{1}{K^2}\E\left[\rvv_k\rvv_k^{\top} - 2\rho\rvv_k\rvv_k^{\top} + \rho^2 \rmI_d\right] \\[3pt]
    &\stackrel{(c)}{=} \frac{1}{K^2}\left(\rho - \rho^2\right) \rmI_d \\[3pt]
    &\stackrel{(d)}{\preceq} \frac{\rho}{K^2}\rmI_d
\end{aligned}
\end{equation}
where $(b)$ comes from the fact that $\rvv_k^{\top}\rvv_k \leq 1$ and $(c)$ is due to the fact that $\E\left[\rvv_k\rvv_k^{\top}\right] = \rho \rmI_d$.

As a result, by introducing the results above with a triangle inequality, we have
\begin{equation}
% \begin{aligned}
    \left\|\sum_{k=1}^K \E\left[\rmX_k^2\right]\right\| \leq \sum_{k=1}^K 
 \left\|\frac{\rho}{K^2}\rmI_d\right\| \leq \frac{\rho}{K} \ .
% \end{aligned}
\end{equation}

% \begin{equation}
% \begin{aligned}
%     \E\left[\left\| \frac{1}{K}\rmV\rmV^{\top} - \rho\rmI_d\right\|\right] 
%     &= \E\left[\left\|\frac{1}{K}\sum_{k=1}^K \left(\rvv_k\rvv_k^{\top} - \rho \rmI_d\right)\right\|\right] \\
%     &= \E\left[\left\|\sum_{k=1}^K \rmX_k \right\|\right] \\
%     &\leq \sqrt{\frac{2\rho\ln(2d)}{K}} + \frac{\ln(2d)}{K}
% \end{aligned}
% \end{equation}

By introducing the results above into Lemma.~\ref{le:Bernstein}, 
% given the fact that $\rho$ is monotonically increasing and $\rho > 1/4$, we have 
\begin{equation}
\begin{aligned}
    \E\left[\left\|\widetilde{\Delta} - \Delta\right\|\right] &= \E\left[\left\|\frac{1}{\rho K}\rmV\rmV^{\top} \Delta - \Delta\right\|\right] \\
    &\leq \E\left[\left\|\frac{1}{\rho K}\rmV\rmV^{\top}  - \rmI_d\right\|\right]\left\|\Delta\right\| \\
    &= \frac{1}{\rho}\E\left[\left\|\sum_{k=1}^K \rmX_k \right\|\right]\left\|\Delta\right\| \\
    &\leq \sqrt{\frac{2\ln(2d)}{\rho K}} + \frac{\ln(2d)}{\rho K} \ ,
    % &\leq \sqrt{\frac{8\ln(2d)}{K}} + \frac{4\ln(2d)}{K}
\end{aligned}
\end{equation}
which finally concludes our proof.

\begin{remark}
\normalfont
Sampling from a truncated normal distribution (rather than a standard normal distribution) ensures a bounded norm, which is crucial for achieving a bounded reconstruction error by our method in Sec.~\ref{sec:proj-update}.
\end{remark}

\subsection{Proof of Thm.~\ref{thm:norm}}\label{appx:proof-norm}

Since the loss function $\ell(\cdot;\cdot)$ is assumed to be $\beta$-smooth w.r.t its first argument, we then have
\begin{equation}
% \begin{aligned}
\ell(\rvw + \epsilon \rvv_k; \rvx^{(i)}) - \ell(\rvw; \rvx^{(i)}) \leq \epsilon \left(\nabla \ell(\rvw; \rvx^{(i)})\right)^{\top} \rvv_k + \frac{1}{2}\beta\epsilon^2\left\|\rvv_k\right\|^2 \leq \epsilon \left(\nabla \ell(\rvw; \rvx^{(i)})\right)^{\top} \rvv_k + \frac{1}{2}\beta\epsilon^2 \ .
% \end{aligned}
\end{equation}

By dividing $\epsilon$ on both sides of the inequality above, we have
\begin{equation}
\begin{aligned}
g_k - \rvv_k^{\top}\nabla \ell(\rvw; \rvx^{(i)})  \leq \frac{1}{2}\beta\epsilon \ . \label{eq:temp-1}
\end{aligned}
\end{equation}

We therefore can conclude our proof using the results below:
\begin{equation}
\begin{aligned}
\left\|\frac{1}{K}\rmV\vg - \frac{1}{K}\rmV\rmV^{\top} \nabla \ell(\rvw; \rvx^{(i)})\right\| &\stackrel{(a)}{=} \left\|\frac{1}{K}\sum_{k=1}^K \left(\rvv_k g_k - \rvv_k\rvv_k^{\top} \nabla \ell(\rvw; \rvx^{(i)})\right)\right\| \\
&\stackrel{(b)}{\leq} \frac{1}{K}\sum_{k=1}^K \left\|\rvv_k g_k - \rvv_k\rvv_k^{\top} \nabla \ell(\rvw; \rvx^{(i)})\right\| \\
&\stackrel{(c)}{\leq} \frac{1}{K}\sum_{k=1}^K \left|g_k - \rvv_k^{\top} \nabla \ell(\rvw; \rvx^{(i)})\right|\left\|\rvv_k\right\| \\[5pt]
&\stackrel{(d)}{\leq} \frac{1}{2}\beta \epsilon
\end{aligned}
\end{equation}
where $(b)$ comes from triangle inequality and $(d)$ results from \eqref{eq:temp-1} and $\left\|\rvv_k\right\| \leq 1$.

\begin{remark}
\normalfont
When $\epsilon \rightarrow 0$, \eqref{eq:temp-1} indicates that $g_k = \rvv_k^{\top}\nabla \ell(\rvw; \rvx^{(i)})$, implying that this scalar gradient in zeroth-order method, e.g., FedKSeed \citep{fedkseed}, is an approximation of directional derivative, i.e., our projected update in \eqref{eq:approx} when $\Delta$ is replaced with $\nabla \ell(\rvw; \rvx^{(i)})$.
\end{remark}

\subsection{Proof of Prop.~\ref{thm:block-wise-speedup}}\label{appx:proof-block-wise-speedup}
Due to the fact that $d = \sum_{l \in [L]} d_l$, $K = \sum_{l \in [L]} K_l$, and $K_l > 0$ for any $l \in [L]$, we have
\begin{equation*}
% \begin{aligned}
 dK = \left(\sum_{l\in [L]} d_l\right) \left(\sum_{l\in [L]} K_l\right) = \sum_{l\in [L]} d_l\left(\sum_{l\in [L]} K_l\right) > \sum_{l\in [L]} d_l K_l \ ,
% \end{aligned}
\end{equation*}
which therefore concludes our proof.

\subsection{Proof of Prop.~\ref{thm:block-wise-error}}\label{appx:proof-block-wise-error}
Based on our block-wise reconstruction in \eqref{eq:block-reconstruct} and Thm.~\ref{thm:reconstruct-error}, we have
\begin{equation}
\begin{aligned}
\E\left[\left\|\widetilde{\Delta} - \Delta\right\|\right] &\stackrel{(a)}{=} \E\left[\sqrt{\sum_{l \in [L]}\left\|\widetilde{\Delta}_l - \Delta_l\right\|^2}\right] \\
&\stackrel{(b)}{<} \E\left[\sqrt{\Bigg(\sum_{l \in [L]}\left\|\widetilde{\Delta}_l - \Delta_l\right\|\Bigg)^2}\right] \\[7pt]
&\stackrel{(c)}{=} \sum_{l \in [L]} \E\left[\left\|\widetilde{\Delta}_l - \Delta_l\right\|\right] \\
&\stackrel{(d)}{\leq} \sum_{l \in [L]} 
 \left(\sqrt{\frac{2\ln(2d_l)}{\rho_l K_l}} + \frac{\ln(2d_l)}{\rho_l K_l}\right) \left\|\Delta_l\right\| \label{eq:temp-2}
\end{aligned}
\end{equation}
where $(a)$ is based on the definition of $\widetilde{\Delta}_l$ and $\Delta_l$ and $(b)$ is from the fact that $\left\|\widetilde{\Delta}_l - \Delta_l\right\| > 0$.

Given that $\sqrt{d_l} > K_l$ and we can then use $\widetilde{\gO}$ to hide the logarithm term in the result above, the following then holds:
\begin{equation}
\begin{aligned}
\E\left[\left\|\widetilde{\Delta} - \Delta\right\|\right] < \widetilde{\gO}\left(\sum_{l \in [L]} 
 \frac{\left\|\Delta_l\right\|}{\rho_l K_l}\right) \ . \label{eq:csnc}
\end{aligned}
\end{equation}

To minimize the upper bound above w.r.t $\{K_l\}_{l=1}^L$ with $\sum_{l \in [L]}K_l= K$, we resort to KKT conditions. Specifically, define $\vk \triangleq [K_1, \cdots, K_L]^{\top}$ and the following Lagrangian function based on $\lambda >0$:
\begin{equation}
F(\vk, \lambda) \triangleq \sum_{l \in [L]} 
 \frac{\left\|\Delta_l\right\|}{\rho_l K_l} + \lambda \left(\sum_{l\in[L]}K_l - K\right) \ .
\end{equation}

To minimize \eqref{eq:csnc}, for any $l \in [L]$, $K_l$ and $\lambda$ then needs to satisfy the following condition:
\begin{equation}
\begin{aligned}
\frac{\partial F(\vk, \lambda)}{\partial K_l} = -\frac{\left\|\Delta_l\right\| / \rho_l}{K_l^2} + \lambda = 0 \ .
\end{aligned}
\end{equation}

That is,
\begin{equation}
\lambda = \frac{\left\|\Delta_1\right\| / \rho_1}{K_1^2} = \cdots = \frac{\left\|\Delta_L\right\| / \rho_L}{K_L^2} \ .
\end{equation}
This finally leads to $K_l \propto \sqrt{\left\|\Delta_L\right\| / \rho_L}$, which consequently concludes our proof.

\begin{remark}
\normalfont
Prop.~\ref{thm:block-wise-error} provides a looser bound than Thm.~\ref{thm:reconstruct-error}, primarily owing to the inequality $(b)$ in \eqref{eq:temp-2}. Based on this looser bound, one might expect that block-wise reconstruction would incur a larger error compared to the vanilla reconstruction in \eqref{eq:reconstruct}. However, empirical results in Appx.~\ref{appx:ablation-reconstruct} and Appx.~\ref{appx:ablation-converge} show that block-wise reconstruction yields comparable performance to the vanilla approach.
\end{remark}

% To prove this result is strictly worse than the one in Thm.~\ref{thm:reconstruct-error}, we aim to show that
% \begin{equation}
% \begin{aligned}
% \sum_{l \in [L]} 
%  \frac{\left\|\Delta_l\right\|}{\rho_l K_l} > \frac{\left\|\Delta\right\|}{\rho K} \ .
% \end{aligned}
% \end{equation}

% Specifically, we 

\subsection{Proof of Thm.~\ref{thm:convergence}}\label{appx:proof-convergence}
Of note, we follow the general idea in \citep{shu2024heterogeneous} to prove the convergence of \ours{}. To begin with, we introduce the following lemmas borrowed from \citep{shu2024heterogeneous}:
\begin{lemma}\label{le:triangle}
Let $\left\{\vu_{1}, \ldots, \vu_{\tau}\right\}$ be any $\tau$ vectors in $\mathbb{R}^{d}$. Then the following holds for any $a>0$:
\begin{align}
    \left\|\vu_i\right\|\left\|\vu_j\right\| &\leq \frac{a}{2}\left\|\vu_i\right\|^2 + \frac{1}{2a}\left\|\vu_j\right\|^2 \ , \label{eq:triangle-1} \\
    \left\|\vu_{i}+\vu_{j}\right\|^{2} &\leq (1+a)\left\|\vu_{i}\right\|^{2}+\left(1+\frac{1}{a}\right)\left\|\vu_{j}\right\|^{2} \ , \label{eq:triangle-2} \\
    \left\|\sum_{i=1}^{\tau} \vu_{i}\right\|^{2} &\leq \tau \sum_{i=1}^{\tau}\left\|\vu_{i}\right\|^{2} \ . \label{eq:triangle-3}
\end{align}
\end{lemma}

\begin{lemma}\label{le:contractive}
For any $\beta$-smooth function $f$, inputs $\vx, \vy$ in the domain of $f$, the following holds for any constant $\eta>0$:
\begin{equation*}
\|\vx-\eta \nabla f(\vx)-\vy+\eta \nabla f(\vy)\|^{2} \leq (1+\eta\beta)^2\|\vx-\vy\|^{2} \ . \label{eq:contractive-1}
\end{equation*}
\end{lemma}

% According to our problem setup in Sec.~\ref{sec:setup} and Algo.~\ref{alg:fetus}, the following naturally holds:
% \begin{equation}
% \begin{aligned}
% \rvw_{r+1} - \rvw_{r} &= - \frac{ \eta}{\rho KN} \rmV_r\rmV_r^{\top}\sum_{i=1}^N \sum_{t=1}^{T} \nabla \ell(\rvw^{(i)}_{r,t-1};\rvx^{(i)}_{r, t-1}) \ , \\
% \E\left[\rvw_{r+1} - \rvw_{r}\right] &= - \frac{ \eta}{N} \sum_{i=1}^N\sum_{t=1}^T \nabla \gL^{(i)}(\rvw^{(i)}_{r,t-1}) \ .
% \end{aligned}
% \end{equation}

Let $\eta \leq 1/(T\beta)$, we can bound the discrepancy between $\rvw_{r, t}^{(i)}$ and $\rvw_r$ for any client $i$ as below
\begin{equation}
\begin{aligned}
&\E\left[\left\|\rvw_{r, t}^{(i)} - \rvw_r\right\|^2\right] \\
% \stackrel{(a)}{=}\ & \eta^2\E\left[\left\|\sum_{\tau=1}^t \nabla \ell(\rvw_{r,\tau-1}^{(i)}; \rvx_{r,\tau-1}^{(i)})\right\|^2\right] \\
\stackrel{(a)}{=}\ & \E\left[\left\|\rvw_{r, t-1}^{(i)} - \eta \nabla \ell(\rvw_{r,t-1}^{(i)}; \rvx_{r,t-1}^{(i)}) - \rvw_r \right\|^2\right] \\[5pt]
\stackrel{(b)}{=}\ & \E\left[\left\|\rvw_{r, t-1}^{(i)} - \eta \nabla \gL(\rvw_{r,t-1}^{(i)}) + \eta \nabla \gL(\rvw_r) - \rvw_r + \eta\left(\nabla \gL(\rvw_{r,t-1}^{(i)}) - \nabla \gL^{(i)}(\rvw_{r,t-1}^{(i)})\right) \right.\right.\\
&\qquad \qquad \left.\left. + \eta\left(\nabla \gL^{(i)}(\rvw_{r,t-1}^{(i)}) - \nabla \ell(\rvw_{r,t-1}^{(i)}; \rvx_{r,t-1}^{(i)}) - \nabla \gL(\rvw_r)\right)\right\|^2\right] \\
\stackrel{(c)}{\leq}\ & \frac{T}{T-1}\E\left[\left\|\rvw_{r, t-1}^{(i)} - \eta \nabla \gL(\rvw_{r,t-1}^{(i)}) + \eta \nabla \gL(\rvw_r) - \rvw_r\right\|^2\right] \\
&\qquad \qquad + 2\eta^2T\,  \E\left[\left\|\nabla \gL^{(i)}(\rvw_{r,t-1}^{(i)}) - \nabla \ell(\rvw_{r,t-1}^{(i)}; \rvx_{r,t-1}^{(i)})\right\|^2 + \left\|\nabla \gL(\rvw_r)\right\|^2\right] \\
\stackrel{(d)}{\leq}&\ \frac{T(1 + \eta \beta)^2}{T-1} \E\left[\left\|\rvw_{r, t-1}^{(i)} - \rvw_r\right\|^2\right] + 2\eta^2T \sigma^2 + 2\eta^2 T\,\left\|\nabla \gL(\rvw_r)\right\|^2 \\[8pt]
\stackrel{(e)}{\leq}&\  24 \eta^2 T^2 \sigma^2 + 24 \eta^2 T^2 \left\|\nabla \gL(\rvw_r)\right\|^2 \label{eq:temp-5}
\end{aligned}
\end{equation}
where $(a)$ is from the local update of $\rvw_{r,t-1}^{\smash{(i)}}$ on each client $i$, and $(c)$ is based on \eqref{eq:triangle-2} in Lemma~\ref{le:triangle} with $a = 1/(T-1)$ and $\gL(\rvw_{r,t-1}^{(i)}) = \gL^{(i)}(\rvw_{r,t-1}^{(i)})$. Besides, $(d)$ results from Lemma~\ref{le:contractive} and the assumption that $\E[\|\nabla \gL^{\smash{(i)}}(\rvw) - \nabla\ell(\rvw;\rvx)\|^2] \leq \sigma^2$. Finally, $(e)$ comes from the summation of geometric series and the fact that $\eta\beta \leq 1/T$ as well as
\begin{equation}
\begin{aligned}
    \sum_{\tau=0}^{t-1} \left(\frac{(T+1)^2}{T(T-1)}\right)^{\tau} &\leq \sum_{\tau=0}^{T-1} \left(\frac{(T+1)^2}{T(T-1)}\right)^{\tau} \\
    &= \frac{\left((T+1)^2/[T(T-1)]\right)^T - 1}{(T+1)^2/[T(T-1)] - 1} \\
    &= \frac{T(T-1)}{3T+1}\left(\left(1 + \frac{3T+1}{T(T-1)}\right)^T - 1\right) \\
    &< \frac{T(T-1)}{3T+1}\left(\exp\left(\frac{3T+1}{T}\right) - 1\right) \\
    &< \frac{T}{3}\left(\exp\left(\frac{7}{2}\right) - 1\right) \\[7pt]
    &< 12T \ . \label{eq:inter-3}
\end{aligned}
\end{equation}

Besides $\E\left[\rmV_r\rmV_r^{\top}\right] = \rho\rmI_d$, one can also verify that $\E\left[\rmV_r\rmV_r^{\top}\rmV_r\rmV_r^{\top}\right] = \rho^2 \rmI_d$, we therefore have
\begin{equation}
\begin{aligned}
&\E\left[\left\|\rvw_{r+1} - \rvw_{r}\right\|^2\right] \\[5pt]
=\ & \E\left[\left(\rvw_{r+1} - \rvw_{r}\right)^{\top}\left(\rvw_{r+1} - \rvw_{r}\right)\right] \\[4pt]
=\ & \E\left[\left(\frac{ \eta}{\rho KN}\right)^2 \left(\sum_{i=1}^N \sum_{t=1}^{T} \nabla \ell(\rvw^{(i)}_{r,t-1};\rvx^{(i)}_{r, t-1})\right)^{\top} \rmV_r\rmV_r^{\top}\rmV_r\rmV_r^{\top}\sum_{i=1}^N \sum_{t=1}^{T} \nabla \ell(\rvw^{(i)}_{r,t-1};\rvx^{(i)}_{r, t-1})\right] \\
=\ & \left(\frac{ \eta}{\rho KN}\right)^2 \E\left[\left(\sum_{i=1}^N \sum_{t=1}^{T} \nabla \ell(\rvw^{(i)}_{r,t-1};\rvx^{(i)}_{r, t-1})\right)^{\top} \E\left[\rmV_r\rmV_r^{\top}\rmV_r\rmV_r^{\top}\right]\sum_{i=1}^N \sum_{t=1}^{T} \nabla \ell(\rvw^{(i)}_{r,t-1};\rvx^{(i)}_{r, t-1})\right] \\
=\ & \left(\frac{ \eta}{N}\right)^2\E\left[\left\|\sum_{i=1}^N \sum_{t=1}^{T} \nabla \ell(\rvw^{(i)}_{r,t-1};\rvx^{(i)}_{r, t-1})\right\|^2\right] \\
\leq\ & \frac{\eta^2}{N}\sum_{i=1}^N \E\left[\left\|\sum_{t=1}^{T}  \nabla \ell(\rvw^{(i)}_{r,t-1};\rvx^{(i)}_{r, t-1})\right\|^2\right]
% \leq\ & 24 \eta^2 T^2 G + 24 \eta^2 T^2 \left\|\nabla \gL(\rvw_r)\right\|^2
\end{aligned}
\end{equation}
where the last inequality comes from the \eqref{eq:triangle-3} in Lemma~\ref{le:triangle}. Here, we omit the subscript $r$ from the random bases $\rmV$ in our notation for simplicity.

Since $\E\left[\left\|\rvw_{r, t}^{(i)} - \rvw_r\right\|^2\right] = \eta^2\E\left[\left\|\sum_{\tau=1}^t \nabla \ell(\rvw_{r,\tau-1}^{(i)}; \rvx_{r,\tau-1}^{(i)})\right\|^2\right]$, by replacing $\tau$ with $T$, we have
\begin{equation}
\E\left[\left\|\rvw_{r+1} - \rvw_{r}\right\|^2\right] \leq 24 \eta^2 T^2 \sigma^2 + 24 \eta^2 T^2 \left\|\nabla \gL(\rvw_r)\right\|^2 \ . \label{eq:temp-3}
\end{equation}

Besides, since $\E\left[\rvw_{r+1} - \rvw_{r}\right] = - \frac{ \eta}{N} \sum_{i=1}^N\sum_{t=1}^T \nabla \gL^{(i)}(\rvw^{(i)}_{r,t-1})$, we have
\begin{equation}
\begin{aligned}
& \E\left[\nabla \gL(\rvw_{r})^{\top}(\rvw_{r+1} - \rvw_{r})\right] \\
\stackrel{(a)}{=}\ & - \frac{ \eta}{N} \E\left[\sum_{i=1}^N\sum_{t=1}^T \nabla \gL(\rvw_{r})^{\top} \nabla \gL^{(i)}(\rvw^{(i)}_{r,t-1})\right] \\
\stackrel{(b)}{=}\ & - \frac{ \eta}{N} \E\left[\sum_{i=1}^N\sum_{t=1}^T \nabla \gL(\rvw_{r})^{\top} \left(\nabla \gL^{(i)}(\rvw^{(i)}_{r,t-1}) - \nabla \gL(\rvw^{(i)}_{r,t-1}) + \nabla \gL(\rvw^{(i)}_{r,t-1}) - \nabla \gL(\rvw_r) + \nabla \gL(\rvw_r)\right)\right] \\
\stackrel{(c)}{\leq}\ & \frac{ \eta}{N} \sum_{i=1}^N\sum_{t=1}^T\left(\eta\beta T\left\|\nabla \gL(\rvw_r)\right\|^2 + \frac{\beta}{4\eta T}\E\left[\left\|\rvw^{(i)}_{r,t-1} - \rvw_r\right\|^2\right]\right) - \eta T \left\|\nabla \gL(\rvw_r)\right\|^2 \\[7pt]
\stackrel{(d)}{\leq}\ & (7\eta^2T^2\beta - \eta T)\left\|\nabla \gL(\rvw_r)\right\|^2 + 6\eta^2 T^2\beta \sigma^2
\end{aligned}
\end{equation}
where $(c)$ comes from Cauchy–Schwarz inequality and the fact that $\gL(\rvw_{r,t-1}^{(i)}) = \gL^{(i)}(\rvw_{r,t-1}^{(i)})$. In addition, $(d)$ results from \eqref{eq:temp-3}.

Finally, based on the assumption that $\gL$  is $\beta$-smooth, we naturally have
\begin{equation}
\begin{aligned}
\E\left[\gL(\rvw_{r+1}) - \gL(\rvw_{r})\right] &\leq \E\left[\nabla \gL(\rvw_{r})^{\top}\left(\rvw_{r+1} - \rvw_{r}\right)\right] + \frac{\beta}{2}\E\left[\left\|\rvw_{r+1} - \rvw_{r}\right\|^2\right] \\[8pt]
&\leq (19\eta^2T^2\beta - \eta T)\E\left[\left\|\nabla \gL(\rvw_r)\right\|^2\right]  + 18\eta^2 T^2\beta \sigma^2 \ .
% &\leq (72\eta^4T^4\beta^3+ 21\eta^2T^2\beta - \eta T)\left\|\nabla \gL(\rvw_r)\right\|^2 \\
% &\qquad\qquad + \left(18\eta^4T^4\beta^3 + 20\eta^2T^2\beta\right)(G+D) + \frac{1}{2\beta}D
\end{aligned}
\end{equation}

By rearranging and letting $\eta \leq \frac{1}{20 T\beta}$, we have
\begin{equation}
\begin{aligned}
\E\left[\left\|\nabla \gL(\rvw_r)\right\|^2\right] \leq \frac{20\ \E\left[\gL(\rvw_{r}) - \gL(\rvw_{r+1})\right]}{\eta T} + 360 \eta T\beta \sigma^2 \ ,
\end{aligned}
\end{equation}

Finally, by summarizing both sides over $R$ rounds and scaling them with $1/R$, we have the following results to conclude our proof:
\begin{equation}
\begin{aligned}
\min_{r \in [R)}\E\left[\left\|\nabla \gL(\rvw_r)\right\|^2\right] \leq \frac{20\left(\gL(\rvw_{0}) - \min_{\rvw} \gL(\rvw)\right)}{\eta T R} + 360 \eta\beta T \sigma^2 \ . \label{eq:temp-4}
\end{aligned}
\end{equation}

\begin{remark}
\normalfont
Note that the large constant in \eqref{eq:temp-4} arises from our bound in \eqref{eq:temp-5} for sufficiently large $T$. This bound can be improved in practice by considering a smaller $T$ instead.
\end{remark}

\clearpage
\section{Experiments}\label{appx:exp-sec}
\subsection{Experimental Setup}\label{appx:exp-setup}
\paragraph{Baselines.} In line with the comparison in \citep{fedkseed}, we selected four practical methods for federated LLM tuning as our baselines: (1) FedPTuning \citep{fedptuning}, (2) FedPrompt \citep{FedPrompt}, (3) FedIT \citep{FedIT}, and (4) FedIT-SGD, a variant of FedIT that replaces Adam with SGD. In addition, we included four full-parameter tuning methods for comparison: (1) FedAvg \citep{fedavg}, (2) FedZO \citep{fedzo}, (3) FedMeZO, a hybrid of FedAvg and MeZO \citep{mezo}, and (4) FedKSeed \citep{fedkseed}. 

% In addition, we follow the practice in \citep{fedkseed} to construct datasets and evaluate the performance of LLMs.

\subsubsection{Setup on the Natural Instruction and Dolly-15K Datasets}
\paragraph{Datasets.} We conducted our experiments using the Natural Instructions (NI) \citep{dataset-ni} and Dolly-15K \citep{dataset-dolly} datasets, following a setup similar to \citep{fedkseed}. For the NI dataset, we allocated 738 training tasks to individual clients for local updates and reserved 119 test tasks for global evaluation, reflecting a non-IID distribution. Meanwhile, for the Dolly-15K dataset, the final task was utilized for global evaluation, while the remaining tasks were distributed among 200 clients with varying levels of label distribution skew. Rouge-L \citep{lin2004rouge} was chosen as the evaluation metric. Given our resource constraints, we selected DataJuicer-1.3B \citep{juicer} and LLaMA-3B \citep{llama} as the base models for our study. The corresponding HuggingFace model paths are ``datajuicer/LLaMA-1B-dj-refine-150B" and ``openlm-research/open\_llama\_3b".

% \paragraph{FL Settings.} In each round of federated learning, 5\% of clients were randomly selected to participate. Following the same practice in FedKSeed \citep{fedkseed}, we set the total number of communication rounds to 40 for the NI dataset and 60 for Dolly-15K for all baselines. Due to the compelling efficiency of our method, we set the total number of communication rounds to 12 for the NI dataset and 20 for Dolly-15K for \ours{}. First-order baselines trained locally for one epoch, and FedKSeed trained for 200 steps, while our \ours{} algorithm trained for 10 iterations (i.e., $T=10$ in Algo.~\ref{alg:fetus}). The $K$ value was set to 4096 for FedKSeed. All approaches perform local update with a batchsize of 1 to reduce memory consumption. For each local update iteration in \ours{}, we accumulate the gradients from 4 samples.

\paragraph{Hyper-parameters.} For \ours{}, the local update learning rate $\eta$ for each client is set to $1 \times 10^{-4}$, where the selected learning rate is searched from $[2 \times 10^{-4}, 1 \times 10^{-4}, 5 \times 10^{-5}]$. The global aggregation learning rates on Natural Instruction and Dolly-15K are set to $10.0$ and $3.0$, respectively, which is search from $[10.0, 5.0, 1.0]$. For other baselines in Tab.~\ref{tab:comparison} of our main paper, we reported their accuracy performances using the results from FedKSeed~\citep{fedkseed}.

\paragraph{Prompt Template.}
In our experiments, the raw input data is pre-processed to follow a structured format, where we warp the input text to the Alpaca prompt template \citep{taori2023stanford}. The corresponding templates for the NI and Dolly-15K dataset are shown in Tab.~\ref{app:tab:template_NI} and \ref{app:tab:template_Dolly}.

\begin{table}[ht]
\caption{Prompt template for Natural Instructions.}
\begin{tcolorbox}[boxrule=0.5pt]
\begin{tabular}{p{\textwidth}}
Below is an instruction that describes a task, paired with an input that provides further context. Write a response that appropriately completes the request. \\ \\
\#\#\# Instruction: \{{\fontfamily{ccr}\selectfont \textcolor{blue}{Definition}}\} \\ \\
\#\#\# Input: \{{\fontfamily{ccr}\selectfont \textcolor{blue}{input}}\} \\ \\
\#\#\# Response:
\end{tabular}
\end{tcolorbox}
\label{app:tab:template_NI}
\end{table}

\begin{table}[ht]
\caption{Prompt template for Dolly-15K. If some data instances do not have the {\fontfamily{ccr}\selectfont context} attribute, we will discard the line ``\#\#\# Input: '' in the template.}
\begin{tcolorbox}[boxrule=0.5pt]
\begin{tabular}{p{\textwidth}}
Below is an instruction that describes a task, paired with an input that provides further context. Write a response that appropriately completes the request. \\ \\
\#\#\# Instruction: \{{\fontfamily{ccr}\selectfont \textcolor{blue}{instruction}}\} \\ \\
\#\#\# Input: \{{\fontfamily{ccr}\selectfont \textcolor{blue}{context}}\} \\ \\
\#\#\# Response:
\end{tabular}
\end{tcolorbox}
\label{app:tab:template_Dolly}
\end{table}

\subsubsection{Setup on the CodeAlpaca and GSM8K Datasets}
\paragraph{Datasets.} To further demonstrate that \ours{} can also improve the capability of larger LLMs for code generation and mathematical reasoning, we conducted more experiments using the CodeAlpaca~\citep{chaudhary2023code} and GSM8K~\citep{cobbe2021training} datasets, following a similar federated setup. 
The CodeAlpaca dataset (of around 8.0k samples) is a code dataset that consists of ten programming languages, including C, C\#, C++, Go, Java, PHP, Pascal, Python, Scale, and X86-64 Assemble. We exclude the X86-64 Assembly data due to limited samples in the dataset. We uniformly randomly sampled 10\% instances from the original data as the hold-out test set for evaluation, and we split the remaining 10\% samples into nine subsets based on the programming language category and assign each subset to one client as its local training data. For GSM8K, its official train set is split into three subsets, where each client’s dataset consists of grade school math questions randomly partitioned from the original dataset, forming a IID distribution. We use the official GSM8K test split as the evaluation dataset. Rouge-L \citep{lin2004rouge} was chosen as the evaluation metric. To demonstrate the scalability of \ours{}, we extended the experiments to larger models: LLaMA2-7B and LLaMA2-13B \citep{llama} as the base models for our study. The corresponding HuggingFace model paths are ``meta-llama/Llama-2-7b-hf" and ``meta-llama/Llama-2-13b-hf".

% \paragraph{FL Settings.} Due to the computing constraints, we set the total number of communication rounds to 20 for both CodeAlpaca and GSM8K for all methods. The $K$ value was set to 4096 for FedKSeed as the same as before. Zeroth-order baselines are trained locally for 200 steps, while FedAvg and \ours{} are trained for 10 iterations with accumulating the gradients from 4 samples. All approaches perform local updates with a batch size of 1 to reduce memory consumption.

\paragraph{Hyperparameters.} For FedZO and FedKSeed, the local update learning rate is set to $3\times 10^{-7}$ for all models. For FedAvg on both LLaMA2-7B and LLaMA2-13B, the local update learning rate $\eta$ for each client is set to $3 \times 10^{-4}$, and the global aggregation learning rate is set to $1.0$. For \ours{} on LLaMA2-7B, the local update learning rate $\eta$ is set to $3 \times 10^{-4}$ and the global aggregation learning rate is set to $5.0$. For \ours{} on LLaMA2-13B, the local update learning rate $\eta$ is set to $5 \times 10^{-4}$ and the global aggregation learning rate is set to $10.0$. The selected learning rate is searched from $[5 \times 10^{-4}, 3 \times 10^{-4}, 1 \times 10^{-4}]$ and the selected global aggregation learning rates is searched from $[10.0, 5.0, 1.0]$.

\subsection{Calculation of Computational Cost and Communication Overhead}\label{appx:comp&comm}

In this subsection, we provide the details of how the computational cost and communication cost are calculated for all methods listed in Tab.~\ref{tab:computation:cost}.

\paragraph{Calculation of  Computational Cost.} For FedZO, we follow the same hyper-parameters ($b_1=200, b_2=1$) for FedZO from FedKSeed paper~\citep{fedkseed}, which employs 200 local update steps and 1 perturbation for each local update step. For calculating the computational cost of FedAvg and \ours{}, we apply 10 local update steps for each client. Same as our experimental setting in Tab.~\ref{tab:comparison}, the batch size is set to 1 for all methods. The time cost incurred at gradient projection is also included in the Local Update.

In the global aggregation process for both FedZO and FedAvg, raw gradients from all clients are averaged and then used to update the global model. In contrast, for FedKSeed and \ours{}, the projected gradients are first aggregated through averaging, then reconstructed, and finally used to update the global model.

For the overall computation cost per round, we follow the calculation below:
\[
\text{Overall} = \text{Local Update} + \text{Global Aggr.}
\]

\paragraph{Calculation of Communication Overhead.} The per-round communication cost refers to the total number of parameters exchanged between a client and the central server during a single round. This includes both the raw or projected gradients that the client sends to the server and the aggregated gradients that the client receives from the server. Each parameter (or projected gradient) is encoded as 16-bit floating point numbers. In accordance with the practice in FedKSeed, we set the number of rounds $R$ to 40 for both FedZO and FedKSeed. Given the notable convergence rate of \ours{}, we set $R$ to 12 for both \ours{} and FedAvg. Although \citep{fedkseed} employs $R = 40$ for FedAvg, we use $R = 12$ to provide a strong basis for comparison and to highlight the computational efficiency of \ours{}.

% For the overall communication cost, we follow the calculation below:
% \[
% \text{Overall} = \text{Per-Round} \times R.
% \]

\subsection{More Comparison on LLaMA3-8B and Qwen2.5-7B Models}\label{appx:more-comparison-qwen}
We conducted additional experiments on CodeAlpaca and GSM8K using LLaMA3-8B and Qwen2.5-7B in Tab.~\ref{tab:rouge_l_comparisons_qwen}. The results demonstrate the consistent effectiveness of our \ours{}, achieving near FedAvg performance with significantly reduced communication overhead across models and tasks.

\begin{table}[t]
\centering
\setlength{\cmidrulewidth}{0.5pt}
\caption{More comparison of Rouge-L (\%) among various algorithms on LLaMA3-8B and Qwen2.5-7B models. Each cell reports the mean $\pm$ std of Rouge-L scores from the final round of four runs, each with a different random seed. The highest and second-highest scores are shown in \textbf{bold} and \underline{underline}, respectively.
}
\label{tab:rouge_l_comparisons_qwen}
% \resizebox{0.485\textwidth}{!}{
\begin{tabular}{lcccc}
\toprule
\multirow{2}{*}{\textbf{Algorithm}} & \multicolumn{2}{c}{\textbf{CodeAlpaca}} & \multicolumn{2}{c}{\textbf{GSM8K}} \\
\cmidrule(l){2-3} \cmidrule(l){4-5}
& LLaMA3-8B & Qwen2.5-7B & LLaMA3-8B & Qwen2.5-7B \\ 
\midrule
FedZO & 16.66 $\pm$ 0.50 & 9.14 $\pm$ 0.32 & 7.79 $\pm$ 1.36 & 23.84 $\pm$ 1.19 \\
FedKSeed & 5.73 $\pm$ 1.26 & 9.14 $\pm$ 0.32 & 7.79 $\pm$ 1.36 & 23.84 $\pm$ 1.19  \\
FedAvg & \textbf{19.88} $\pm$ 0.67 & \textbf{17.47} $\pm$ 0.49 & \textbf{45.48} $\pm$ 0.51 & \textbf{43.86} $\pm$ 0.36 \\
\midrule
\ours{} (ours) &  \underline{19.59} $\pm$ 0.66 & \underline{14.64} $\pm$ 0.74 & \underline{45.07} $\pm$ 0.78 & \underline{38.28} $\pm$ 1.70\\
\bottomrule
\end{tabular}
% }
\vspace{-2mm}
\end{table}

\subsection{More Comparison of Computational Cost and Communication Overhead}\label{appx:more-comparison}

Tab.~\ref{tab:computation:cost:13b_gsm8k} compares the computational cost and communication overhead of LLaMA2-13B using the GSM8K dataset. Because of GPU memory constraints, FedZO and FedAvg have slightly higher computational costs, as gradients need to be stored on the CPU. The results show that even for large models like LLaMA2-13B, \ours{} still demonstrates superior scalability. Compared to FedKSeed, \ours{} reduces computational costs significantly: 9.8$\times$ for local updates, 3.9$\times$ for global aggregation, and 4.5$\times$ for overall cost per bound. Additionally, compared to FedAvg, which does not utilize shared randomness, \ours{} achieves a dramatic 10$^7\times$ reduction in communication costs. These results, along with the evidence in Sec.~\ref{sec:exp-results}, further highlight the scalability of \ours{} in federated full-parameter tuning.

\begin{table}[t]
\centering
\setlength{\cmidrulewidth}{0.5pt}
\caption{Comparison of computational cost and communication overhead on LLaMA2-13B, focusing on (a) the computational costs from local updates, global aggregation, and the overall tuning process; and (b) the per-round and overall communication costs. The improvement achieved by our \ours{} is reported in brackets using \textcolor{blue}{blue} (compared with FedKSeed) and \textcolor{orange}{orange} (compared with FedAvg).}
% \resizebox{\textwidth}{!}{
\begin{tabular}{lcccc}
\toprule
\multirow{2}{*}{\textbf{Algorithm}}
& \multicolumn{3}{c}{\textbf{Computational Cost} (Sec.)} & \multicolumn{1}{c}{\textbf{Communication Cost}} \\
\cmidrule(l){2-4}   
 & Local Update & Global Aggr. & Overall & (\# params.) \\
\midrule
\cellcolor{gray!20} FedZO  & \cellcolor{gray!20} 114.1 & \cellcolor{gray!20} 25.7 & \cellcolor{gray!20} 139.8 & \cellcolor{gray!20} 2.6$\times$10$^{10}$  \\
\cellcolor{blue!20} FedKSeed  &\cellcolor{blue!20} 188.4 & \cellcolor{blue!20} 666.2 &\cellcolor{blue!20}  854.6 &\cellcolor{blue!20} 8.2$\times$10$^3$  \\
 \cellcolor{orange!20} FedAvg  &  \cellcolor{orange!20} 24.9 &  \cellcolor{orange!20} 25.7 & \cellcolor{orange!20}  50.6  &  \cellcolor{orange!20} 2.6$\times$10$^{10}$ \\
\midrule
\ours{} (ours) & \textbf{19.2} \textcolor{blue}{(9.8$\times$)} & \textbf{169.4} \textcolor{blue}{(3.9$\times$)} & \textbf{188.6} \textcolor{blue}{(4.5$\times$)} & \textbf{7.6$\times$10$^3$} \textcolor{orange}{(10$^6\times$)} \\
\bottomrule
\end{tabular}
% }
\label{tab:computation:cost:13b_gsm8k}
\end{table}

\subsection{Ablation Studies on Reconstruction}\label{appx:ablation-reconstruct}

\begin{figure}[ht]
\raisebox{3mm}{\begin{minipage}{0.5\textwidth}
\centering
\hspace{-6mm}
\includegraphics[width=0.94\textwidth]{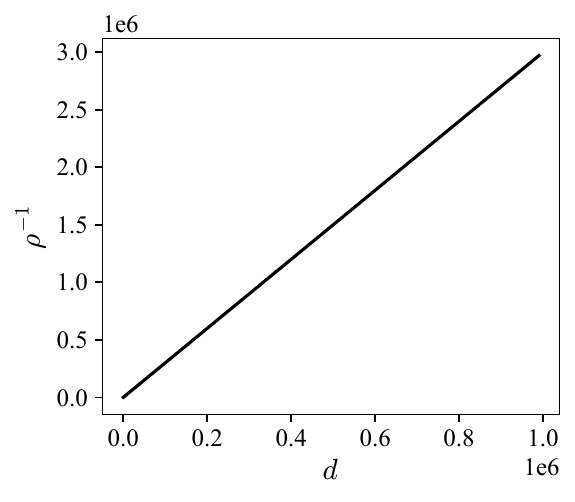}
\captionsetup{width=0.99\textwidth}
\vspace{-4mm}
\captionof{figure}{Rate of $1 / \rho$ w.r.t. dimension $d$.}
\label{fig:rho}
\end{minipage}}
\begin{minipage}{0.5\textwidth}
\centering
\hspace{-4mm}
\includegraphics[width=0.95\textwidth]{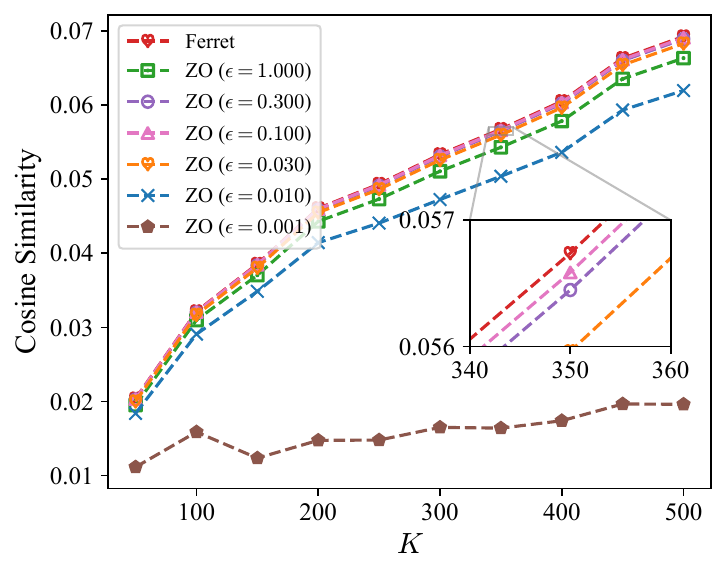}
\vspace{-2mm}
\captionsetup{width=0.98\textwidth}
\captionof{figure}{Reconstruction accuracy of our \eqref{eq:block-reconstruct} vs. zeroth-order method under varying $K$ and $\epsilon$.}
\label{fig:reconstruct_K}
\end{minipage}
% \vspace{-3mm}
\end{figure}

\paragraph{Rate of $1/\rho$ w.r.t Dimension $d$.} In Fig.~\ref{fig:rho}, we present the rate of $1/\rho$ where $\rho$ is defined in Sec.~\ref{sec:proj-update} to verify our claim following Thm.~\ref{thm:reconstruct-error}. The results in Fig.~\ref{fig:rho} confirm that $1/\rho$ indeed follows a rate of $\gO(d)$.

\paragraph{Comparison of Reconstruction Accuracy between \ours{} and ZO Method.} In Fig.~\ref{fig:reconstruct_K}, we present the reconstruction accuracy (measured by cosine similarity) for the $d=10^5$-dimensional gradient of the function $F(\vx) = \sum_{i=1}^d x^2_i$ at a randomly sampled input $\vx$ with varying $K$ by using our method in \eqref{eq:reconstruct} and zeroth-order method with different values of $\epsilon$. The goal is to compare the reconstruction accuracy of our \eqref{eq:reconstruct} with that of the ZO method under varying $K$ and $\epsilon$. The results in Fig.~\ref{fig:reconstruct_K} indicate that: \textit{(a)} our method \eqref{eq:reconstruct} achieves improved reconstruction accuracy compared to the ZO method, particularly the one with an optimal $\epsilon=0.1$, which indeed aligns with the insights from our Thm.~\ref{thm:norm}; \textit{(b)} both our method \eqref{eq:reconstruct} and the ZO method exhibit the same increasing rate in reconstruction accuracy as $K$ increases, highlighting the connection between these two methods as implied by our Thm.~\ref{thm:norm}; and \textit{(c)} this increasing rate is generally linear, which is consistent with Thm.~\ref{thm:reconstruct-error}. These results therefore further verify the insights in Thm.~\ref{thm:reconstruct-error} and Thm.~\ref{thm:norm}, and support the advantages of our method \eqref{eq:reconstruct} over the ZO method.

\begin{figure}[t]
\vspace{-2mm}
\centering
\begin{tabular}{ccc}
    \hspace{-4.5mm}
    \includegraphics[width=0.317\columnwidth]{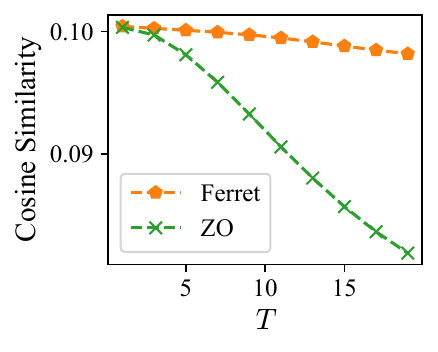}&
    \hspace{-4.5mm}
    \includegraphics[width=0.334\columnwidth]{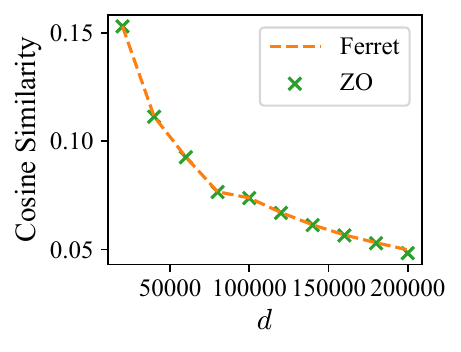}&
    \hspace{-6mm}
    \raisebox{-0.75mm}{\includegraphics[width=0.398\columnwidth]{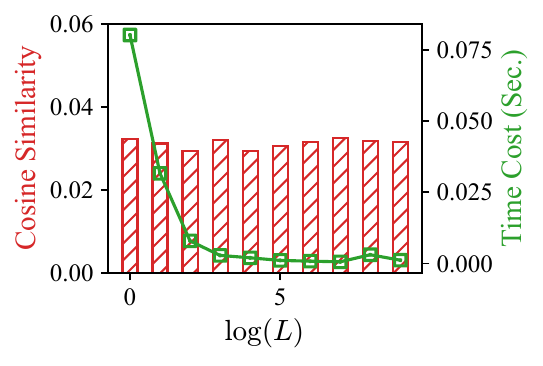}} \\
    {\hspace{1.5mm}(a) Varying $T$} & {(b) Varying $d$} & {\hspace{-10mm} (c) Varying $L$}
\end{tabular}
\caption{
Reconstruction Accuracy (measured by cosine similarity between reconstruction and ground truth) of our \eqref{eq:block-reconstruct} vs. zeroth-order method under varying $T$, $d$, and $L$. 
}
\label{fig:ablation-reconstruct}
% \vspace{-5mm}
\end{figure}

\paragraph{Reconstruction Accuracy of \ours{} under Varying $T$.} In Fig.~\ref{fig:ablation-reconstruct} (a), we present the reconstruction accuracy (measured by cosine similarity) of a $T$-iteration gradient descent update for the function $F(\vx) = \sum_{i=1}^d \sin^2(x_i)$ with a learning rate of 0.1, $d=5\times10^4$, $L=1$, and $K=500$, using our method in \eqref{eq:block-reconstruct} and the zeroth-order (ZO) method described in Thm.~\ref{thm:norm} with $\epsilon=0.1$. The goal is to compare the accumulated error from our \eqref{eq:block-reconstruct} with that of the ZO method. Interestingly, Fig.~\ref{fig:ablation-reconstruct} (a) shows that our method maintains consistent reconstruction accuracy as the number $T$ of gradient descent iterations increases, whereas the ZO method experiences a noticeable decline in accuracy. This result implies that our \eqref{eq:block-reconstruct} effectively avoids the accumulated error typical in zeroth-order methods, aligning with the theoretical justification provided in Sec.~\ref{sec:thm-reconstruct}.

\paragraph{Reconstruction Accuracy of \ours{} under Varying $d$.} In Fig.~\ref{fig:ablation-reconstruct} (b), we show the reconstruction accuracy (measured by cosine similarity) of $d$-dimensional gradient of $F(\vx) = \sum_{i=1}^d \sin^2(x_i)$ at a randomly sampled input $\vx$, with $L=1$ and $K=500$, using our method in \eqref{eq:block-reconstruct} and the zeroth-order (ZO) method described in Thm.~\ref{thm:norm} with $\epsilon=0.1$. The goal is to compare the reconstruction accuracy rate with respect to the dimension $d$ between our \eqref{eq:block-reconstruct} method and the ZO method. Interestingly, Fig.~\ref{fig:ablation-reconstruct} (b) shows that both methods achieve the same reconstruction accuracy rate with respect to $d$. More importantly, when $d$ becomes large, the accuracy rate is approximately linear, which aligns with the theoretical insights provided in Thm.~\ref{thm:reconstruct-error}.

\paragraph{Reconstruction Accuracy of \ours{} under Varying $L$.} In Fig.~\ref{fig:ablation-reconstruct} (c), we present the reconstruction accuracy (measured by cosine similarity) and computational complexity (measured by time cost) for the $d=5.12\times10^5$-dimensional gradient of function $F(\vx) = \sum_{i=1}^d \sin^2(x_i)$ at a randomly sampled input $\vx$, under varying $L$ of the same number of dimensions and $K=512$, using our method in \eqref{eq:block-reconstruct}. The goal is to study the impact of block size $L$ on our \eqref{eq:block-reconstruct}. Notably, Fig.~\ref{fig:ablation-reconstruct} (c) shows that our block-wise reconstruction \eqref{eq:block-reconstruct} significantly reduces computational complexity (in line with Prop.~\ref{thm:block-wise-speedup}), while maintaining consistent reconstruction accuracy as $L$ increases. These results further verify the efficacy of our block-wise reconstruction \eqref{eq:block-reconstruct}.

\subsection{More Ablation Studies on Convergence and Generalization}\label{appx:ablation-converge}

\begin{figure}[t]
\vspace{-2mm}
\centering
\begin{tabular}{cc}
    \hspace{-12mm}
    \includegraphics[width=0.365\columnwidth]{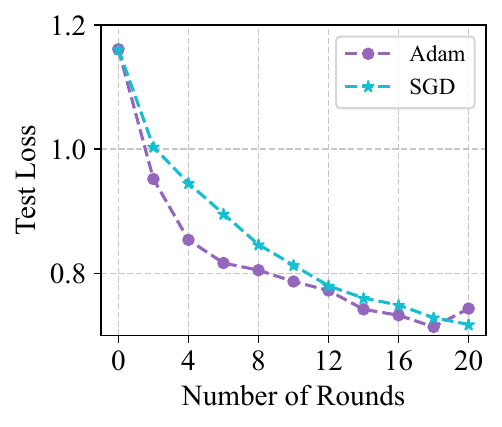}&
    \hspace{3.2mm}
    \raisebox{4.0mm}{
    \includegraphics[width=0.348\columnwidth]{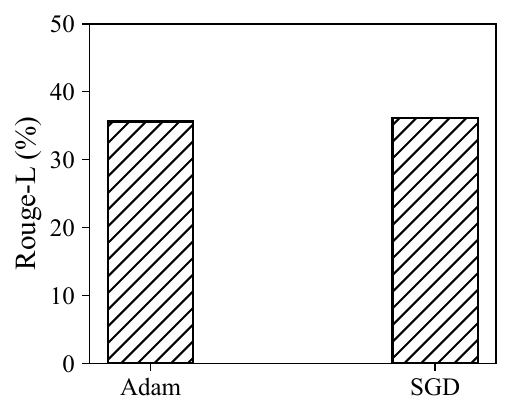}} \\
    {\hspace{-10mm} (a) Convergence} & {\hspace{7mm}(b) Generalization}
\end{tabular}
\caption{
Convergence and generalization of \ours{} under varying optimizers for local updates. 
}
\label{fig:opt}
% \vspace{-5mm}
\end{figure}

\paragraph{Convergence and Generalization of \ours{} under Varying Optimizers.} In Fig.~\ref{fig:opt}, we present the convergence and generalization of \ours{} under different optimizers for its local updates, using the same experimental setup described in Appx.~\ref{appx:exp-setup}. Notably, Fig.~\ref{fig:opt} demonstrates that \ours{} achieves faster convergence with an improved optimizer (e.g., Adam vs. SGD) while maintaining comparable generalization performance. These findings further support the adaptability of \ours{}, as discussed in Sec.~\ref{sec:scale}.

\end{appendices}

\end{document}